\documentclass[sn-basic]{sn-jnl}
\usepackage{times}
\usepackage{CJKutf8}
\usepackage{epstopdf}
\usepackage{graphicx}%
\usepackage{multirow}%
\usepackage{amsmath,amssymb,amsfonts}%
\usepackage{amsthm}%
\usepackage{mathrsfs}%
\usepackage[title]{appendix}%
\usepackage{xcolor}%
\usepackage{textcomp}%
\usepackage{manyfoot}%
\usepackage{booktabs}%
\usepackage{subcaption}
\usepackage{footnote}
\usepackage{caption}
\usepackage{algorithm}%
\usepackage{algorithmicx}%
\usepackage{algpseudocode}%
\usepackage{listings}%
\usepackage{algpseudocode}
\usepackage{float}
\usepackage{geometry}
\newtheorem{definition}{Definition}%
\floatname{algorithm}{Algorithm} %算法
\begin{document}
\begin{CJK}{UTF8}{gbsn}

\title[Article Title]{An Adaptive Cost-Sensitive Learning and Recursive Denoising Framework for Imbalanced SVM Classification}

\author[1]{\fnm{Lu} \sur{Jiang}}\email{LL99\_J@163.com}
\author*[1,2]{\fnm{Qi}\sur{Wang}}\email{wangqimath@mail.neu.edu.cn}
\author[1]{\fnm{Yuhang}\sur{Chang}}\email{damonchang53@gmail.com}
\author[3]{\fnm{Jianing}\sur{Song}}\email{s1374520908@163.com}
\author[1]{\fnm{Haoyue}\sur{Fu}}\email{fuhaoyue@mail.neu.edu.cn}
\author[4,5]{\fnm{Xiaochun}\sur{Yang}}\email{yangxc@mail.neu.edu.cn}

\affil[1]{\orgdiv{College of Sciences}, \orgname{Northeastern University}, \orgaddress{ \city{Shenyang}, \postcode{110819}, \country{People's Republic of China}}}

\affil[2]{\orgdiv{Key Laboratory of Data Analytics and Optimization for Smart Industry}, \orgname{ Northeastern University}, \orgaddress{\city{Shenyang}, \postcode{110819}, \country{People's Republic of China}}}

\affil[3]{\orgdiv{School of Mechanical Engineering and Automation}, \orgname{Northeastern University}, \orgaddress{\city{Shenyang}, \postcode{110819}, \country{People's Republic of China}}}

\affil[4]{\orgdiv{Software College}, \orgname{Northeastern University}, \orgaddress{\city{Shenyang}, \postcode{110819}, \country{People's Republic of China}}}

\affil[5]{\orgdiv{School of Computer Science and Engineering}, \orgname{Northeastern University}, \orgaddress{\city{Shenyang}, \postcode{110819}, \country{People's Republic of China}}}

\abstract{Category imbalance is one of the most popular and important issues in the domain of classification. Emotion classification model trained on imbalanced datasets easily leads to unreliable prediction. The traditional machine learning method tends to favor the majority class, which leads to the lack of minority class information in the model. Moreover, most existing models will produce abnormal sensitivity issues or performance degradation. We propose a robust learning algorithm based on adaptive cost-sensitivity and recursive denoising, which is a generalized framework and can be incorporated into most stochastic optimization algorithms. The proposed method uses the dynamic kernel distance optimization model between the sample and the decision boundary, which makes full use of the sample's prior information. In addition, we also put forward an effective method to filter noise, the main idea of which is to judge the noise by finding the nearest neighbors of the minority class. In order to evaluate the strength of the proposed method, we not only carry out experiments on standard datasets but also apply it to emotional classification problems with different imbalance rates (IR). Experimental results show that the proposed general framework is superior to traditional methods in Accuracy, G-mean, Recall and F1-score.}
\keywords{Imbalanced data · Robust learner · Adaptive Cost-Sensitive Learning · Prior information · Recursive Denoising}

\maketitle
\section{Introduction}\label{}
Support Vector Machine (SVM), a typical statistical learning method, was first proposed by Vapnik and Cortes in 1995 \cite{1}. SVM has gained significant popularity as a classification and regression algorithm. It has been used in various real-world problems widely and successfully, including particle recognition, text classification \cite{2,3,4}, visual recognition \cite{5}, feature selection\cite{6,7}, and bioinformatics \cite{8}. SVM encounters various obstacles, including the incorporation of previous knowledge from the datasets \cite{9,10} and the identification of a suitable kernel function for mapping the datasets to a separable space of high dimensions \cite{11}, and expediting the resolution of quadratic programming (QP) problems \cite{12,13,14}. Our exclusive focus lies in utilizing SVM to address the challenges posed by large-scale imbalanced datasets. In doing so, we employ the technique of assigning distinct weights to individual samples. The SVM classification hyperplane is uniquely determined by support vectors (SVs). A sample that satisfies $y_{i}f(x_{i})=1$ is called a boundary support vector. The two planes that are parallel to the optimal classification hyperplane and intersect the boundary support vector are referred to as classification interval hyperplanes.

\begin{equation}
\begin{aligned}
&H_{1}:f(x_{i})=1,y_{i}=+1\\
&H_{2}:f(x_{i})=-1,y_{i}=-1
\end{aligned}
\end{equation}

The error support vector lies between the hyperplanes $H_{1}$ and $H_{2}$ in the classification interval. Once the optimal classification hyperplane is acquired, the decision function can be expressed as
\begin{equation}
    f(x)=\text{sgn}(w^{T}x+b)
\end{equation}
where $\text{sgn}(\cdot)$ represents the sign function.

Noise frequently contaminates the training data in several practical engineering applications. Furthermore, certain data points within the training dataset are erroneously positioned into the feature space alongside other categories, which are referred to as outliers. As indicated in \cite{15} and \cite{16}, outliers frequently transform into support vectors that have higher Lagrangian coefficients throughout the training process. The classifier created using the support vector machine exclusively depends on support vectors. The existence of outliers or noise in the training set leads to a large deviation of the decision border from the ideal plane, when using the standard SVM training technique. This high sensitivity of SVM to outliers and noise is a consequence of this deviation. This work proposes the inclusion of weights to showcase the contributions of various samples. 

Our primary focus is on the fundamental problem of soft-margin SVM. Wang et al.\cite{45} improved the soft-margin support vector machine and proposed a new fast support vector machine with truncated square hinge loss, which fully improved the calculation speed of large-scale datasets. Descent techniques can be employed to minimize the objective function of a conventional SVM, which is known for its high convexity. Nevertheless, descent methods necessitate accurate calculation of the gradient of the goal function, a task that is frequently arduous. The Stochastic Gradient Descent (SGD) method addresses this issue by employing unbiased gradient estimates using tiny data samples \cite{25,26}. It is the primary approach for solving large-scale stochastic optimization issues. Nevertheless, the practical attractiveness of SGD approaches remains restricted due to their need for a substantial number of iterations to achieve convergence. Indeed, SGD exhibits sluggish convergence due to its reliance on gradients, a drawback that is further amplified when gradients are substituted with random approximations. Various stochastic optimization strategies have been employed thus far to address SVM models. Aside from first-order techniques such as SGD, second-order approaches are also extensively employed. Second-order stochastic optimization methods impose greater demands on functions compared to first-order methods, as they are capable of quadratic differentiation. Consequently, studying these approaches is a demanding task. Second-order stochastic optimization methods inherently address the issues of limited precision and sluggish convergence that are associated with first-order approaches. This paper considers solving SVM using second-order stochastic optimization methods to further reduce the impact of outliers and dataset imbalance while ensuring accuracy.

In this paper, we introduce a novel generalized framework with adaptive weight function for imbalanced SVM classification (AW-WSVM). Our main contributions are as follows: 

1. This work proposes a novel cost-sensitive learning technique. In the kernel space, the contribution of training samples to the optimal classification hyperplane is determined by their kernel distance from the hyperplane. Specifically, the closer a sample is to the hyperplane, the greater the penalty term for its loss, while the farther a sample is from the hyperplane, the smaller the penalty term for its loss. We suggest implementing a new adaptive weight function that can adjust itself according to the changes in the decision hyperplane.

2. We introduce a new soft-margin weighted support vector machine. The model no longer focuses on the overall loss but instead emphasizes the loss of each sample.

3. We filter the data during each iteration, paying more attention to the samples in the vicinity of the decision hyperplane. As a small number of samples are sensitive to noise, we propose a recursive method to find the nearest neighbors and eliminate noisy data.

4. Employ the adaptive cost-sensitive learning and recursive denoising technique we suggested in conjunction with various optimization algorithms, and conduct experiments using datasets with different characteristics. In the field of emotion classification, we have done many experiments under different imbalanced ratios. The results demonstrate that the suggested weighting technique can enhance the performance of SVM even more.

The remaining sections of this paper are structured in the following manner. Section 2 provides a concise overview of the recent literature on the classification of imbalanced data. Section 3 presents the introduction of a novel adaptive cost-sensitive learning and recursive denoising technique. Section 4 involves conducting experiments on various datasets by integrating different algorithms with our suggested technique. The discussion and conclusions can be found in section 5.
\section{Related Work}\label{}
\subsection{Imbalanced Data}
Data imbalances occur in the real world, often leading to overfitting. Faced with the problem of imbalance in SVM, there are two popular solutions. The first is to reconstruct the training points. Increase the number of points of a few classes in different ways (oversampling) or decrease the number of points of a majority class (undersampling) \cite{37,38}. The second is to introduce appropriate weight coefficients into the loss function. The error cost or the decision threshold of imbalanced data is adjusted to increase the weight of the minority class and reduce the weight of the majority class to achieve balance. An initial approach to incorporating weighting into SVM is through the use of fuzzy support vector machine (FSVM) \cite{17}. Fuzzy SVM assigns different fuzzy membership degrees to all samples in the training set, and retrains the classifier. Shao et al.\cite{41} introduced an extremely efficient weighted Lagrangian twin support vector machine (WLTSVM) that utilizes distinct training points to create two nearby hyperplanes for the classification of unbalanced data. Fan et al. \cite{39} proposed an improved method based on FSVM to determine fuzzy membership by class determinism of samples, so that distinct samples can provide different contributions to the classification hyperplane.Kang et al. \cite{40} introduced a downsampling method for Support Vector Machines (SVM) that uses spatial geometric distances. This method assigns weights to the samples depending on their Euclidean distance from the hyperplane. The weights highlight the samples' influence on the decision hyperplane. In order to improve the performance of SVM to generate samples, paper \cite{44} combines oversampling with particle swarm optimization algorithm to find the best sample position.
\subsection{Outliers of SVM}
In the real world, datasets often possess substantial size and exhibit imbalances, hence posing challenges to the problem-solving process.  Yang et al.\cite{18} proposed Weighted Support Vector Machine (WSVM), which addresses the issue of anomaly sensitivity in SVM by assigning varying weights to different data points. The WSVM approach utilizes the kernel-based Probability c-means (Kpcm) algorithm to provide varying weights to the primary training data points and outliers. Wu et al.\cite{19} introduced the Weighted Margin Support Vector Machine (WMSVM), a method that combines prior information about the samples and gives distinct weights to different samples. Zhang et al. \cite{20} proposed a Density-induced Margin Support Vector Machine (DMSVM), for a given dataset, DMSVM needs to extract the relative densities of all training data points. The densities can be used as the relative edges of the corresponding training data points. This can be considered as an exceptional instance of WMSVM. In \cite{21}, particle swarm optimization (PSO) is used for the time weighting of SVM. Du et al.\cite{22} introduced a fuzzy-compensated multiclass SVM approach to improve the outlier and noise sensitivity issues, proposing how to dynamically and adaptively adjust the weights of the SVM. Zhu et al.\cite{23} introduced a weighting technique based on distance, where the weights are determined by the proximity of the nearest diverse sample sequences, referred to as extended nearest neighbor chains. The paper \cite{24} presents a novel approach called the weighted support vector machine (WSVM-FRS), which involves incorporating weighting coefficients into the penalty term of the optimization problem. Inadequate samples possess minimal weights, while significant samples possess substantial weights.
\subsection{Stochastic Optimization Algorithm for SVM}
Gradient-based optimization techniques are widely used in the development of neural systems and can also be applied to solve SVM. First-order approaches are commonly employed due to their straightforwardness and little computational cost. Some examples of optimization algorithms are stochastic gradient Descent (SGD)\cite{42}, AdaGrad\cite{27}, RMSprop\cite{28}, and Adam\cite{29}. Although the first-order approach is easy to understand, its slow convergence is a significant disadvantage. Utilizing second-order bending information enables the improvement of convergence. The Broyden-Fletcher-Goldfarb-Shanon (BFGS) approach has garnered significant attention and research in the training of neural networks. Simultaneously, an increasing number of scholars are employing the BFGS algorithm to address Support Vector Machine (SVM) problems. However, a significant drawback of the second-order technique is its reliance on substantial processing and memory resources. Implementing quasi-Newtonian methods in a random environment poses significant challenges and has been a subject of ongoing research. The Online BFGS (oBFGS) approach, mentioned \cite{30}, is a stochastic quasi-Newtonian method that has demonstrated early stability. In contrast to BFGS, this method eliminates the need for line search and alters the update of the Hessian matrix. An analysis is conducted on the worldwide convergence of stochastic BFGS, and a novel online L-BFGS technique is introduced \cite{31}. A novel approach is suggested, which involves subsampling Hessian vector products and collecting curvature information periodically and point-by-point. This method can be classified as a stochastic quasi-Newtonian method \cite{32}. In \cite{33}, the authors introduce a stochastic (online) quasi-Newton method that incorporates Nesesterov acceleration gradients. This strategy is applicable in both complete and limited memory formats. The study demonstrates the impact of various momentum rates and batch sizes on performance.

In the following sections, we apply the proposed framework to SGD, oBFGS and oNAQ to do comparative experiments. We pay more attention to these three algorithms, SGD and oBFGS are summarized in Algorithm \ref{alg:1} and Algorithm \ref{alg:2} respectively. ONAQ will be discussed in the third section.
\begin{algorithm}
\caption{SGD Method\cite{42}}\label{alg:1}
\begin{algorithmic}[1] %每行显示行号，1表示每1行进行显示
\Require minibatch $X_{k}$, $k_{max}$ and learning rate $\epsilon$
\Ensure $\mathbf{w}\in \mathbb{R}^{n}$
\State $k\leftarrow1$
\While {$k<k_{max}$}
\State $\nabla F\leftarrow\nabla F(\mathbf{w},X_{k})$
\State $\mathbf{w}\leftarrow \mathbf{w}-\epsilon \nabla F$
\State $k\leftarrow k+1$
\EndWhile    
\end{algorithmic}
\end{algorithm}

\begin{algorithm}
\caption{oBFGS Method\cite{30}}\label{alg:2}
\begin{algorithmic}[1] %每行显示行号，1表示每1行进行显示
\Require minibatch $X_{k}$, $k_{max}$, $\lambda \ge 0$, $H_{k}=\epsilon I$ and $v_{k}=0$
\Ensure $\mathbf{w}\in \mathbb{R}^{n}$
\State $k\leftarrow1$
\While {$k<k_{max}$}
\State $\nabla F_{1}\leftarrow\nabla F(\mathbf{w_{k}},X_{k})$
\State $g_{k}\leftarrow-H_{k}\nabla F(\mathbf{w_{k}},X_{k})$
\State $g_{k}=\frac{g_{k}}{\lVert g_{k}\rVert_{2}}$
\State Determine $\alpha_{k}$ using 
\State \quad\quad\quad $\alpha_{k}=\frac{\tau}{\tau+k}\alpha_{0}$
\State $v_{k+1}\leftarrow \alpha_{k}g_{k}$
\State $\mathbf{w_{k+1}}\leftarrow \mathbf{w_{k}}+v_{k+1}$
\State $\nabla F_{2}\leftarrow \nabla F(\mathbf{w_{k+1}},X_{k})$
\State $s_{k}\leftarrow \mathbf{w_{k+1}}-\mathbf{w_{k}}$
\State $y_{k}\leftarrow\nabla F_{2}-\nabla F_{1}+\lambda s_{k}$
\State Update $\hat{H}_{k}$ using 
\State $$H_{k+1}=\big(I-\frac{s_{k}y_{k}^{T}}{y_{k}^{T}s_{k}}\big)H_{k}\big(I-\frac{y_{k} s_{k}^{T}}{y_{k}^{T}s_{k}}\big)+\frac{s_{k}s_{k}^{T}}{y_{k}^{T}s_{k}}$$
\State $k\leftarrow k+1$
\EndWhile    
\end{algorithmic}
\end{algorithm}

\section{Proposed Method}\label{}
\subsection{Support Vector Machine(SVM)}
We consider a classical binary classification problem. Suppose there is a training set ${(x_{i},y_{i})}_{i=1}^l$, for any $i\in\{ 1,... ,l\}$, $x_{i}\in \mathbb{R}^{n}$ is the sample, $y_{i}\in \{-1,+1\}$ is the label. The purpose of the SVM classifier is to find a function $f(\cdot)$: $x_{i}\to y_{i}$ that separates all the samples. A linear classification function can be expressed as $f(x)=\mathbf{w^{T}}x+b$, where $\mathbf{w}\in \mathbb{R}^{n}$ is a weight vector and $b\in \mathbb{R}$ is a bias term. If $f(x)>0$, $x$ is assigned to class $+1$; Otherwise $x$ is divided into class $-1$.

\begin{figure}[h]
\centering
\includegraphics[width=0.5\textwidth]{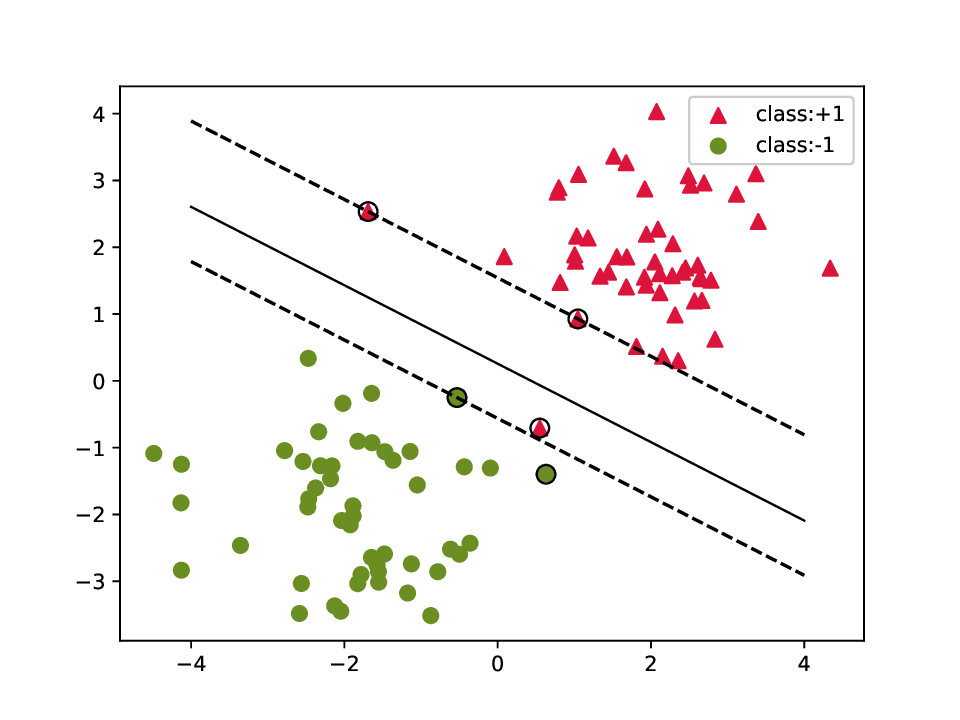}
\caption{Decision hyperplane learned by hard-margin SVMs. Circles are support vectors.}
\label{fig1} 
\end{figure}

The SVM problem is based on the core principle of achieving classification by reducing the error rate and optimizing the hyperplane with the highest margin. This is illustrated in Fig.\ref{fig1}, where the solid line represents the classification hyperplane $f(x)=0$. SVM employs the kernel function $\phi(x_{i})$ to transform low-dimensional features into higher-dimensional features in order to address nonlinear situations. The kernel approach is used to directly calculate the inner product between high-dimensional features, eliminating the requirement for explicit computation of nonlinear feature mappings. Afterwards, linear classification is conducted in the feature space with a high number of dimensions. Soft-margin SVM utilizing kernel techniques can be expressed in the following manner:

\begin{equation}
\begin{aligned}
     \mathop{\min}_{\mathbf{w} \in \mathbb{R}^{n}} F(\mathbf{w})=\cfrac{1}{l} \sum\limits_{i=1}^{l}f(\mathbf{w},\theta_{i})=\cfrac{C}{2}\lVert\mathbf{w}\rVert^{2}+\cfrac{1}{l}\sum\limits_{i=1}^{l}l((\phi(x_{i}),y_{i});\mathbf{w})
\end{aligned}
\end{equation}

where $f(\mathbf{w},\theta_{i} )=\cfrac{C}{2}\lVert\mathbf{w}\rVert^{2}+\text{max}(0,1-y_{i}(\mathbf{w^T}\phi(x_{i})+b))$, and $C>0$ is a hyperparameter. 

\par
\begin{figure}[h]
\centering
\includegraphics[width=0.5\textwidth]{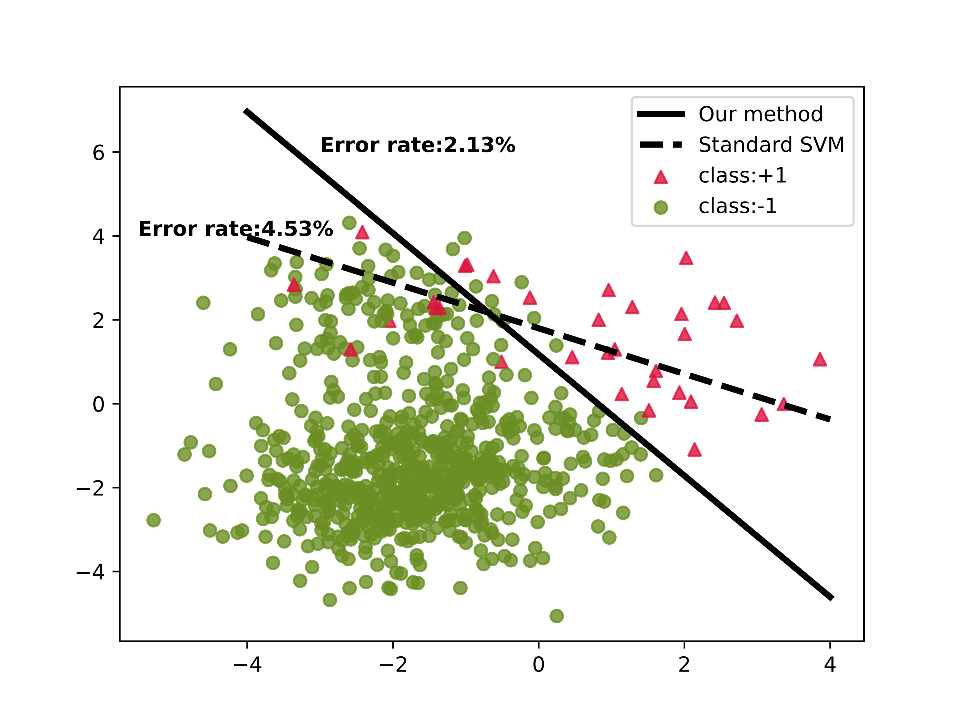}
\caption{Decision hyperplane learned from imbalanced data. The solid line is the hyperplane obtained by the proposed method, while the dashed line is obtained by the standard solving algorithm. }
\label{fig2} 
\end{figure}

 Considering imbalanced datasets, the solution of SVM may appear overfitting. Standard SVMs tend to assign any sample that comes into the class-heavy side. Fig.\ref{fig2} illustrates the above situation. When the dashed line is used as the classification hyperplane, the error rate is as high as 4.53\%. When the solid line is used (the solid line is learned through the method proposed in this paper), the error rate is reduced to 2.13\%. This illustrates the importance of improving data imbalances.

\begin{figure}[h]
\centering
\includegraphics[width=0.5\textwidth]{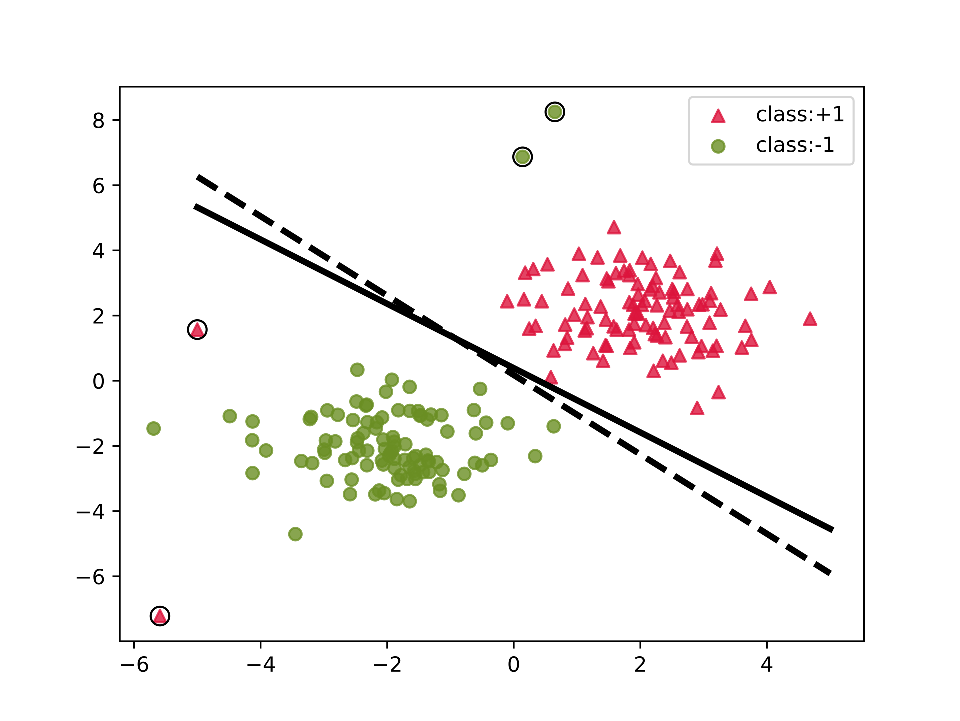}
\caption{Decision hyperplane learned by hard-margin SVMs. The solid line is a decision hyperplane without noise, while the dashed line is a decision hyperplane with noise. The noises are marked as circles.}
\label{fig3} 
\end{figure}

In numerous real-world engineering scenarios, the training data frequently encounters noise or outliers, resulting in substantial deviations in the decision bounds of SVM. The outlier sensitivity problem of ordinary SVM algorithms refers to this phenomena. Fig.\ref{fig3} demonstrates the atypical susceptibility of SVM. In the absence of outliers in the training data, the SVM algorithm can identify the optimal hyperplane, shown by a solid line, that maximizes the margin between the two classes. If an outlier exists on the other side, the decision hyperplane, indicated by a dashed line, deviates considerably from the optimal selection. Consequently, the training procedure of a typical SVM is highly susceptible to outliers.

\subsection{Soft-Margin Weighted Support Vector Machine (WSVM)}
Soft-Margin Weighted Support Vector Machine (WSVM), aims to apply distinct weights to individual data points based on their relative significance within the class. This enables individual data points to exert varying influences on the decision hyperplane. Set $\alpha_{i}$ as the weight of the $ith$ a sample $x_{i}$, then the training data set into $\{(x_{i}, y_{i}, \alpha_{i})\}_{i=1}^{l}$, the $\alpha_{i}$ is continuous and $0\le\alpha_{i}\le1$. Then the optimization problem can be written as:

\begin{equation}\label{eq:4}
\begin{aligned}
    \mathop{\min}_{\mathbf{w} \in \mathbb{R}^{n}} F(\mathbf{w})=\dfrac{1}{l} \sum\limits_{i=1}^{l}f(\mathbf{w},\theta_{i})=\dfrac{C}{2}\lVert\mathbf{w}\rVert^{2}+\dfrac{1}{l}\sum\limits_{i=1}^{l}l((\phi(x_{i}),y_{i},\alpha_{i});\mathbf{w})
\end{aligned}
\end{equation}
where $f(\mathbf{w},\theta_{i} )=\frac{1}{2} C \lVert\mathbf{w}\rVert^{2}+\text{max}(0,\alpha_{i}(1-y_{i}(\mathbf{w^T}\phi(x_{i})+b)))$

We no longer focus on the overall loss, but on the loss of each sample. In other words, we can regard the weight as a penalty term. The greater the punishment for the sample, the less the corresponding loss is required. In fact, the SVM is more accurate and targeted by punishing each sample separately by setting the weight. It satisfies the spacing maximization principle.

In order to make the expression of WSVM more concise, we give the following explanation
\begin{equation}
\begin{aligned}
\textbf{A}&:=[\alpha_{1}, \alpha_{2},...,\alpha_{l}]^{T}\in \mathbb{R}^{l}\\
\textbf{1}&:=[1, 1,...,1]^{T}\in \mathbb{R}^{l}\\
\textbf{y}&:=[y_{1}, y_{2},...,y_{l}]^{T} \in \mathbb{R}^{l}\\
\textbf{Z}&:=[y_{1}\phi(x_{1}), y_{2}\phi(x_{2}),...,y_{l}\phi(x_{l})]^{T}\in \mathbb{R}^{l\times n}\\
\end{aligned}
\end{equation}

Therefore, problem (\ref{eq:4}) can be expressed more concisely as follows

\begin{equation}\label{eq:6}
\begin{aligned}
    \mathop{\min}_{\mathbf{w} \in \mathbb{R}^{n}} F(\mathbf{w})
    &=\frac{C}{2}\lVert\mathbf{w}\rVert^{2}+\frac{1}{l}\textbf{A}^{T}(\textbf{1}-\textbf{Z}\mathbf{w}-b\textbf{y})
\end{aligned}
\end{equation}

Obviously, the standard SVM and the proposed WSVM are different in weight coefficient $\alpha_{i}$. The higher the value of $\alpha_{i}$, the more significant the sample's contribution to the decision hyperplane. In conventional SVM, samples are considered equal, meaning that each sample has an equal impact on the decision hyperplane. However, that is not true. In the real datasets, there are imbalanced samples and outliers. The large number of samples and the presence of outliers will result in the decision hyperplane deviating, which is undesirable. Consequently, it is necessary to diminish or disregard the impact of these samples, while amplifying the significance of the limited categories of samples, support vectors, and accurately classified samples. This is what WSVM does. Both WSVM and SVM ensure maximum spacing, facilitating the differentiation between the two categories of data. Furthermore, the former also prioritizes the analysis of samples that have a greater impact.

\subsection{Adaptive Weight Function(AW Function)}
The weight of sample is developed based on the distance. Suppose that the distance between the sample and the hyperplane is $d$ in the feature space, then $d$ obeys a Gaussian distribution with a mean value of 0. Its probability density function is
\begin{equation}  
f(d)=\frac{1}{\sqrt{2\pi}\sigma}e^{-\frac{d^{2}}{2\sigma^{2}}}
\end{equation}

The initial decision hyperplane $\mathbf{w_{0}}\phi(x)+b=0$ is obtained. The set $\mathbf{d_{i}^{k}}$ is the distance between the $i$ th sample and the initial decision hyperplane, in the $k$ th iteration. In order to summarize the expression of $\mathbf{d_{i}^{k}}$, we give several new definitions.

We can easily get different kinds of clustering point in the kernel space. Set the total number of positive sample population as $\mathbf{n_{+}}$ and its clustering point as $\boldsymbol{\xi_{+}}$. In addition, the total number of negative sample groups is recorded as $\mathbf{n_{-}}$, and the clustering point is recorded as $\boldsymbol{\xi_{-}}$. The expressions of $\boldsymbol{\xi_{+}}$ and $\boldsymbol{\xi_{-}}$ are as follows:
\begin{equation}
\boldsymbol{\xi_{+}}=\frac{1}{n_{+}}\sum\limits_{s=1}^{n_{+}}\phi(x_s)
\end{equation}

\begin{equation}
\boldsymbol{\xi_{-}}=\frac{1}{n_{-}}\sum\limits_{t=1}^{n_{-}}\phi(x_t)
\end{equation}

Then the normal vector $\mathbf{w}$ of the decision hyperplane can be expressed as
\begin{equation}
\mathbf{w}=\boldsymbol{\xi_{+}}-\boldsymbol{\xi_{-}}
\end{equation}

In order to calculate the distance between the sample in the kernel space and the decision hyperplane, we set a point on the hyperplane, which is marked as $\boldsymbol{\xi}=\frac{1}{2}(\boldsymbol{\xi_{+}}+\boldsymbol{\xi_{-}})$. Finally, we give the expression of $\mathbf{d_{i}^{k}}$.

\begin{equation}\label{dik}
\begin{aligned}
 \mathbf{d_{i}^{k}}&=\frac{|(\phi(x_{i})-\boldsymbol{\xi})\mathbf{w_{k}}|}{\lVert\mathbf{w_{k}}\rVert}\\
 &=\frac{|\mathbf{w_{k}}^{T}(\phi(x_{i})-\boldsymbol{\xi})|}{\lVert\mathbf{w_{k}}\rVert}=\frac{|\mathbf{w_{k}}^{T}\phi(x_{i})-\mathbf{w_{k}}^{T}\boldsymbol{\xi}|}{\lVert\mathbf{w_{k}}\rVert}\\
 &=\frac{\big|\big(\boldsymbol{\xi_{+}}-\boldsymbol{\xi_{-}}\big)^{T}\phi(x_{i})-\frac{1}{2}\big(\boldsymbol{\xi_{+}}-\boldsymbol{\xi_{-}}\big)^{T}(\boldsymbol{\xi_{+}}+\boldsymbol{\xi_{-}})\big|}{\lVert\mathbf{w_{k}}\rVert}\\
 &=\frac{\Bigg|\frac{1}{n_{+}}\sum\limits_{s=1}^{n_{+}}k(x_{s},x_{i})-\frac{1}{n_{-}}\sum\limits_{t=1}^{n_{-}}k(x_{t},x_{i})}{\lVert\mathbf{w_{k}}\rVert}-\frac{\frac{1}{2}\Big[\frac{1}{n_{+}^{2}}\sum\limits_{s_{1}=1}^{n_{+}}\sum\limits_{s_{2}=1}^{n_{+}}k(x_{s_{1}},x_{s_{2}})-\frac{1}{n_{-}^{2}}\sum\limits_{t_{1}=1}^{n_{-}}\sum\limits_{t_{2}=1}^{n_{-}}k(x_{t_{1}},x_{t_{2}})\Big]\Bigg|}{\lVert\mathbf{w_{k}}\rVert}
\end{aligned}
\end{equation}

We stress the significance of the distance between the sample and the hyperplane. And according to the above assumption, we can give the definition of Adaptive Weight function(AW function).

\begin{definition}[\textbf{AW function}] In the $k$ th iteration, let $\alpha_{ki}$ be the weight of the $i$ th sample. Set $M=\text{max} \{d_{1}^{k},d_{2}^{k},…,d_{l}^{k}\},m=\text{min}\{d_{1}^{k},d_{2}^{k},…,d_{l}^{k}\}$. Accordingly, the following Adaptive Weight function(AW function) is proposed:
\begin{equation}\label{eq:8}
\begin{aligned}
\boldsymbol{\alpha_{ki}}=\frac{2}{\sqrt{2\pi}\sigma}e^{-\frac{(d_{i}^{k})^{2}}{2\sigma^{2}}}+\frac{1}{M-m} e^{-\frac{d_{i}^{k}}{M-m}}
\end{aligned}
\end{equation}
where $M-m$ can weaken the influence of difference or variation amplitude on distribution to some extent.
\end{definition}

\begin{figure}[h]
\centering
\includegraphics[width=8cm]{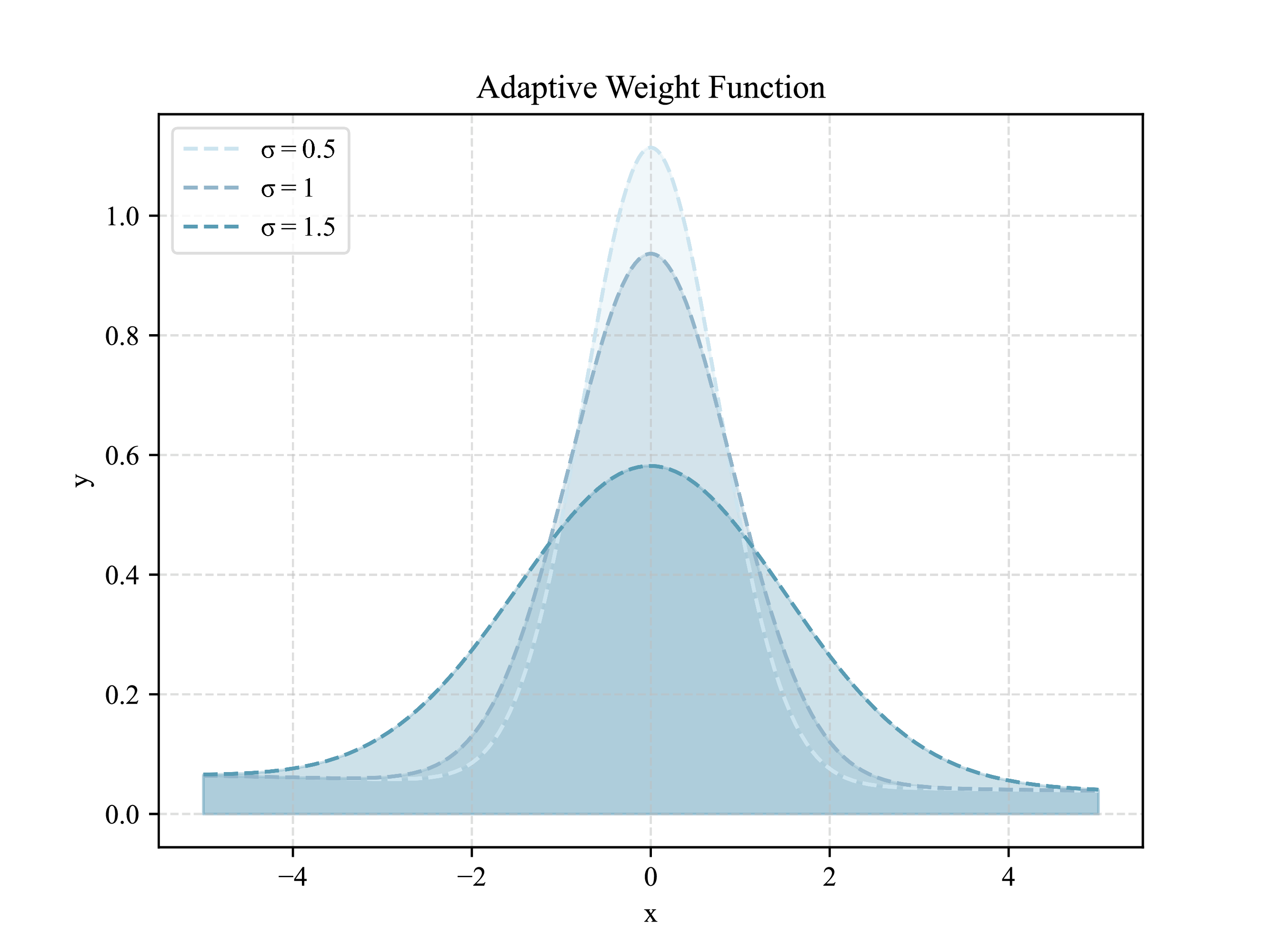}
\caption{AW function with different $\sigma$ ($\sigma\in\{0.5,1,1.5\}$)}
\label{fig4} 
\end{figure}

According to different $\sigma$ values, we draw the AW function image, as shown in Fig.\ref{fig4}. The function exhibits an axis of symmetry at $x=0$ and demonstrates a decreasing trend from the center toward the periphery. In the proposed framework, the x-axis indicates the spatial separation between the sample and the decision hyperplane, while the y-axis signifies the magnitude of the weight assigned to the sample. Therefore, the y-values should be in the range of 0 to 1. It is evident from this analysis that the value of $\sigma$ that yields the greatest match for our weight argument is 1.

Demonstrate the reasonableness of the weighting function for the given set. The weight function should be expressed as a probability density function within each class, meaning that the integral value of the function is equal to 1 within each class. Furthermore, it is important to mention that, to enhance efficiency during the experiment, we will refrain from computing the distance based on the category. Hence, we do not differentiate between the two categories during the process of proof. In brief, we prove that the definite integral of the function equals 2.
\begin{proof}
$$
\int_{0}^{+\infty}\Bigg(\frac{2}{\sqrt{2\pi}\sigma}e^{-\frac{(d_{i}^{k})^{2}}{2\sigma^{2}}}+\frac{1}{M-m} e^{-\frac{d_{i}^{k}}{M-m}}\Bigg)d(d_{i}^{k})
$$
$$
=\int_{0}^{+\infty}\frac{2}{\sqrt{2\pi}\sigma}e^{-\frac{(d_{i}^{k})^{2}}{2\sigma^{2}}}d(d_{i}^{k})\\
+\int_{0}^{+\infty}\frac{1}{M-m} e^{-\frac{d_{i}^{k}}{M-m}}d(d_{i}^{k})
$$
The first term is the probability density function of the Gaussian distribution, so\\
$$ \int_{0}^{+\infty}\frac{2}{\sqrt{2\pi}\sigma}e^{-\frac{(d_{i}^{k})^{2}}{2\sigma^{2}}}d(d_{i}^{k})
$$
$$
=2\int_{0}^{+\infty}\frac{1}{\sqrt{2\pi}\sigma}e^{-\frac{(d_{i}^{k})^{2}}{2\sigma^{2}}}d(d_{i}^{k})
$$
$$
=\int_{-\infty}^{+\infty}\frac{1}{\sqrt{2\pi}\sigma}e^{-\frac{(d_{i}^{k})^{2}}{2\sigma^{2}}}d(d_{i}^{k})\\
=1
$$
The second term is
$$
\int_{0}^{+\infty}\frac{1}{M-m} e^{-\frac{d_{i}^{k}}{M-m}}d(d_{i}^{k})
$$
$$
=\frac{1}{M-m}\int_{0}^{+\infty}e^{-\frac{d_{i}^{k}}{M-m}}d(d_{i}^{k})
$$
$$
=\frac{1}{M-m}\big(-(M-m)e^{-\frac{d_{i}^{k}}{M-m}}\big)\big|_{0}^{+\infty}
$$
$$
=\frac{1}{M-m}(0+M-m)=1
$$
So
$$
\int_{0}^{+\infty}\Bigg(\frac{2}{\sqrt{2\pi}\sigma}e^{-\frac{(d_{i}^{k})^{2}}{2\sigma^{2}}}+\frac{1}{M-m} e^{-\frac{d_{i}^{k}}{M-m}}\Bigg)d(d_{i}^{k})=1+1=2
$$
\end{proof}

For every category of samples, the weight function assigns a weight of 1 when the distance is 0. The weight diminishes proportionally with the distance. It fulfills the condition that samples close to the hyperplane have a greater impact.

\subsection{Generalized Framework for AW-WSVM}
On the basis of the theories in Sections 3.2 and 3.3, we give the most important contribution of this paper, a generalized framework named AW-WSVM. We apply this framework to standard stochastic optimization algorithms, such as Algorithm \ref{alg:1} and Algorithm \ref{alg:2}, to solve WSVM and update the weights with AW function. The standard second-order algorithm oNAQ will be used in the following sections, so it is summarized in Algorithm \ref{alg:3}.

\begin{algorithm}
\caption{oNAQ Method\cite{33}}\label{alg:3}
\begin{algorithmic}[1] %每行显示行号，1表示每1行进行显示
\Require minibatch $X_{k}$, $0<\mu<1$, $k_{max}$, $\hat{H}_{k}=\epsilon I$ and $v_{k}=0$
\Ensure $\mathbf{w}\in \mathbb{R}^{n}$
\State $k\leftarrow1$
\While {$k<k_{max}$}
\State $\nabla F_{1}\leftarrow\nabla F(\mathbf{w_{k}}+\mu v_{k},X_{k})$
\State $\hat{g}_{k}\leftarrow-\hat{H}_{k}\nabla F(\mathbf{w_{k}}+\mu v_{k},X_{k})$
\State $\hat{g}_{k}=\frac{\hat{g}_{k}}{\lVert\hat{g}_{k}\rVert_{2}}$
\State Determine $\alpha_{k}$ using
\State \quad\quad\quad $\alpha_{k}=\frac{\alpha_{0}}{\sqrt{k}}$
\State $v_{k+1}\leftarrow \mu v_{k}+\alpha_{k}\hat{g}_{k}$
\State $\mathbf{w_{k+1}}\leftarrow \mathbf{w_{k}}+v_{k+1}$
\State $\nabla F_{2}\leftarrow \nabla F(\mathbf{w_{k+1}},X_{k})$
\State $p_{k}\leftarrow \mathbf{w_{k+1}}-(\mathbf{w_{k}}+\mu v_{k})$
\State $q_{k}\leftarrow\nabla F_{2}-\nabla F_{1}+\lambda p_{k}$
\State Update $\hat{H}_{k}$ using 
\State $$\hat{H}_{k+1}=\big(I-\frac{p_{k}q_{k}^{T}}{q_{k}^{T}p_{k}}\big)\hat{H}_{k}\big(I-\frac{q_{k} p_{k}^{T}}{q_{k}^{T}p_{k}}\big)+\frac{p_{k}p_{k}^{T}}{q_{k}^{T}p_{k}}$$
\State $k\leftarrow k+1$
\EndWhile    
\end{algorithmic}
\end{algorithm}

All samples are initially assigned the same weight. The assumption in Section 3.3 is given to illustrate the rationality of the above statement. 

Now, we can utilize the AW function mentioned before to produce weights for training the AW-WSVM algorithm. Enter the distance $\mathbf{d_{i}^{k}}(i=1,2,...,l)$ obtained by the $kth$ training into the (\ref{eq:8}). We filter the data by recursively finding the nearest neighbor. Then update the weight of each sample $\alpha_{ki},i=1,...,l$. Finally, we put it in the objective function again. The framework includes the following steps: 

(1) Solve the WSVM to get the hyperplane

Substitute the weight vector obtained from each update into formula (\ref{eq:6}), and get a new hyperplane expression through iterative solution of random optimization algorithm; 

(2) Filter samples

\begin{figure}[h]
\centering
\includegraphics[width=0.5\textwidth]{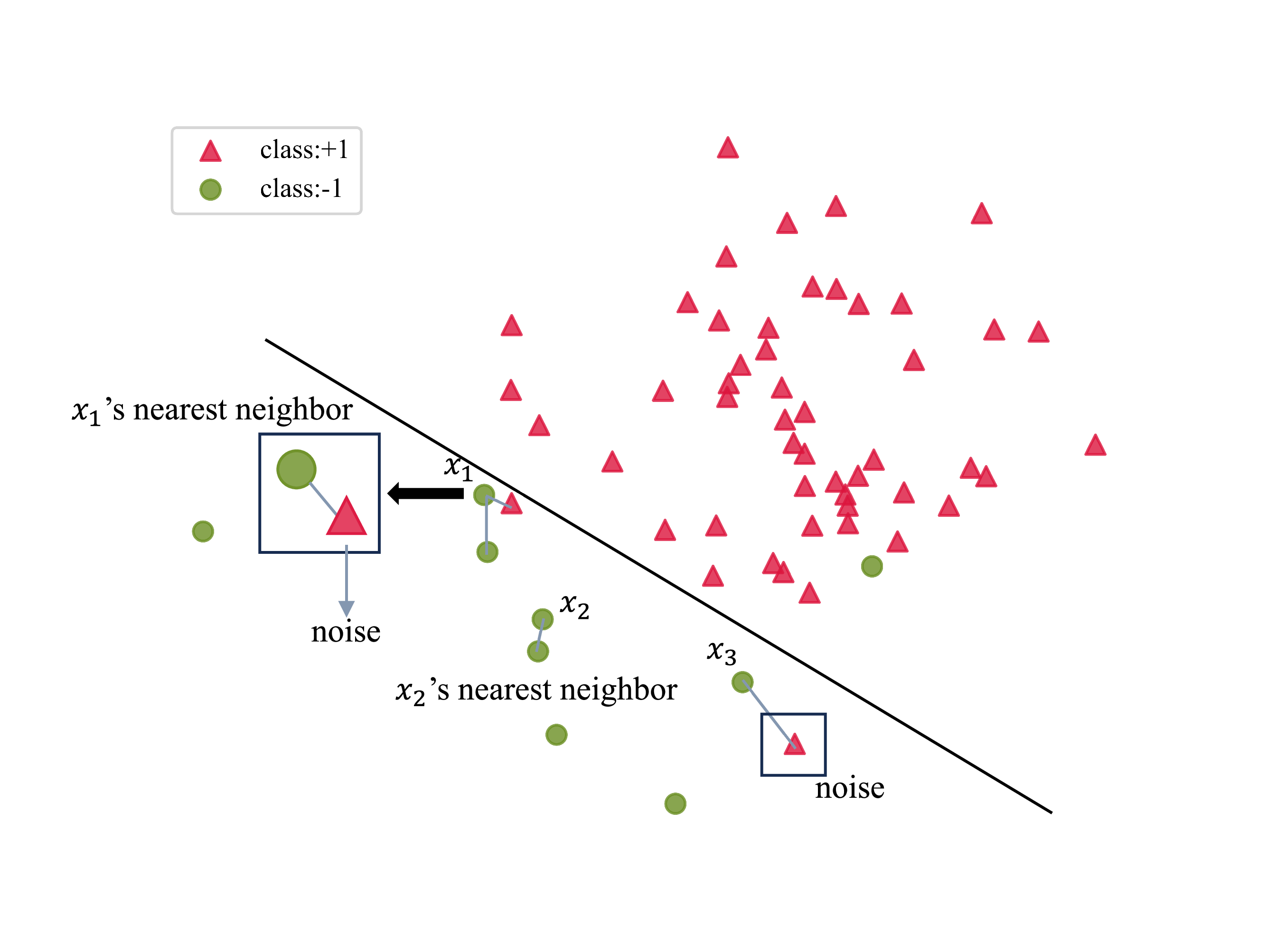}
\caption{The process of filtering samples}
\label{noise} 
\end{figure}

The minority is more sensitive to noise. We can calculate the distance between each sample and the hyperplane, and only pay attention to the samples near the decision boundary (set the threshold according to experience). Because the samples far away from the decision boundary will not have a great influence on the optimal hyperplane. Remember, the majority sample set around the decision hyperplane is $S_{+}$, and the minority sample set is $S_{-}$. Traverse the samples in $S_{-}$ and find the nearest neighbor of $x_{i}$. If the nearest neighbor belongs to $S_{+}$, it means $x_{j}$ is the noise to be removed. We describe this process in detail in Fig. \ref{noise}; 

(3) Update weights

Substitute the obtained sample distance into AW function, and get the weight of each sample.

\begin{figure}[h]
	\centering
	\hfill % 是为了让多幅图在一行均匀分布（不加的效果是都挤在中间）
	\subcaptionbox{Weights of Different Samples\label{fig5(a)}}{
	\includegraphics[width=7.8cm]{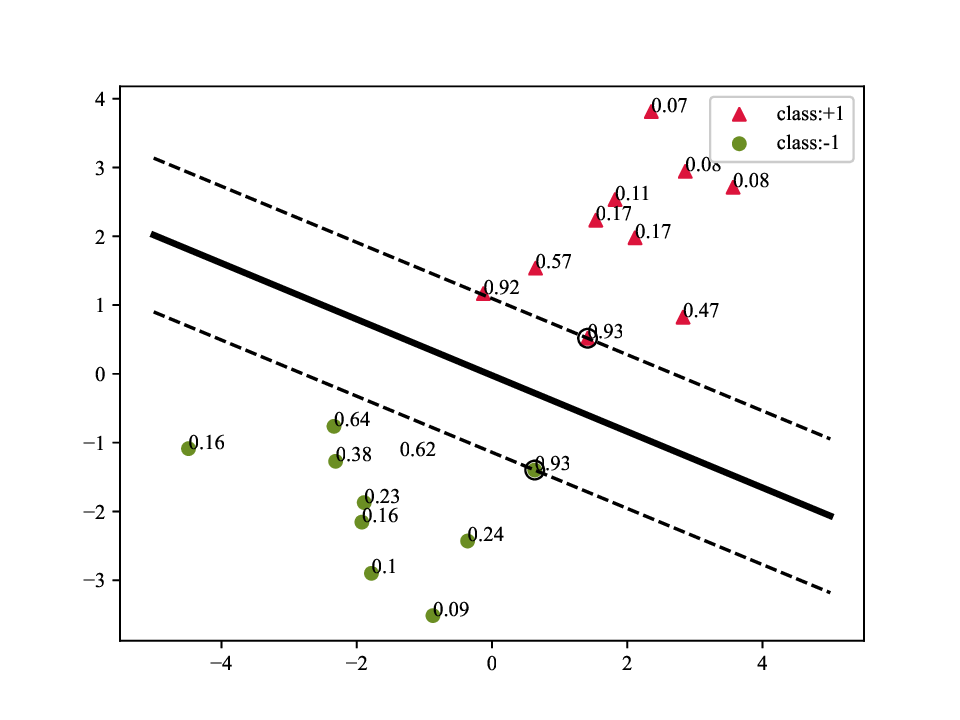}
	}
 \subcaptionbox{3D Visualisation of Weights\label{fig5(b)}}{
		\includegraphics[width=7.8cm]{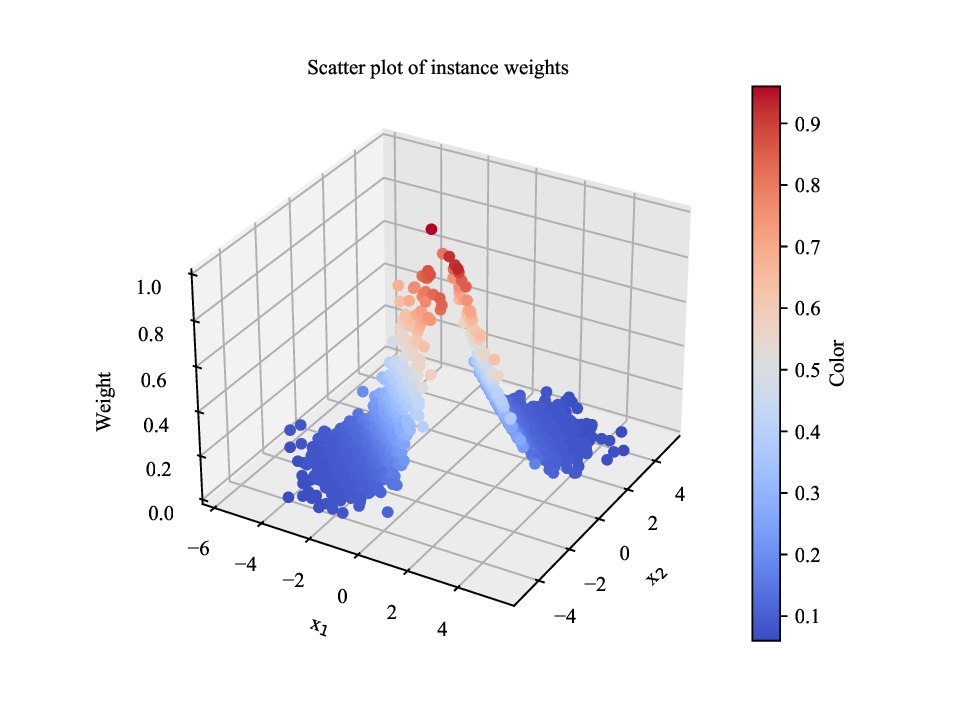}
	}
	\caption{Weights generated by AW function}
	\label{fig5}
\end{figure}

The hyperplane acquired at the $kth$ iteration is used as the beginning input for the $k+1$ iteration. The weight coefficient $\mathbf{w_{k+1}}$ and distance $d_{i}^{k+1}$ of the decision hyperplane are obtained again. Proceed with the aforementioned instructions to iterate. In this way, the outlier will have minimal weight, approaching zero, thereby greatly diminishing its impact on the choice hyperplane. Fig.\ref{fig5} displays the weights obtained for a certain class with the AW function. The data that was used for training points located in the middle region of the data group possess a substantially greater weight (close to 1), while the outliers have a much lower weight (close to 0). The proposed AW-WSVM is summarized in Algorithm \ref{alg:4}. 

\begin{algorithm}
\caption{Proposed AW-WSVM}\label{alg:4} %标题
\begin{algorithmic}[1] %每行显示行号，1表示每1行进行显示
\Require The training set $(x_{i},y_{i},\alpha_{ki}),i=1,…,l$, the initial weight $\alpha_{0}$, the original $\mathbf{w_{0}}=[0,…,0]^{T}$ and $k_{max}$.
\Ensure $\mathbf{w}\in \mathbb{R}^{n}$
\State $k\leftarrow1$
\While {$k<k_{max}$}
\State Obtain a solution $\mathbf{w_{k}}$ by Algorithm \ref{alg:3}
\State Calculate $d_{i}^{k}$ by (\ref{dik})
\State Set $S_{+}$ and $S_{-}$
\For{each $x_{i}\in S_{-}$}
\State Calculate the nearest neighbor of $x_{i}$ and express it as $x_{j}$
\If {$x_{j}\in S_{+}$}
\State Eliminate noise
\EndIf
\EndFor
\State Set
$$ M=\text{max}\{d_{1}^{k},d_{2}^{k},…,d_{l}^{k} \}$$
$$ m=\text{min}\{d_{1}^{k},d_{2}^{k},…,d_{l}^{k}\}$$
\State Update $\alpha_{k}=[\alpha_{k1},…,\alpha_{kl}]^{T}$ using (\ref{eq:8})
\State $k\leftarrow k+1$
\EndWhile    
\end{algorithmic}
\end{algorithm}

We can draw a conclusion that the normal vector of the decision hyperplane in each iteration can be solved by most stochastic optimization algorithms such as Algorithm \ref{alg:1}, \ref{alg:2} or \ref{alg:3}. In the case of Algorithm \ref{alg:3} oNAQ, we can easily provide important steps:
\begin{equation}
  \nabla F_{1}\leftarrow \nabla F(\mathbf{w_{k}}+\mu v_{k},X_{k}) = C(\mathbf{w_{k}}+\mu v_{k})-\frac{1}{l}Z^{T}A_{k} 
\end{equation}

\begin{equation}
  \nabla F_{2}\leftarrow \nabla F(\mathbf{w_{k+1}},X_{k}) = C\mathbf{w_{k+1}}-\frac{1}{l}Z^{T}A_{k+1} 
\end{equation}

\section{Numerical Experiments}\label{}
In this section, we present a comparison between the proposed method AW-WSVM and typical first-order and second-order approaches. The comparison is conducted on different datasets. We use experimental data to illustrate the effectiveness of the proposed framework in solving the problem of imbalanced data classification. These experiments were performed on python3 on MacOS with the Apple M2 chip.

To evaluate the proposed method comprehensively, it is insufficient to solely rely on a single indicator. Before providing the assessment criteria, it is necessary to establish the confusion matrix for the binary classifier. The confusion matrix provides fundamental details on the classifier's actual and expected classifications (see Table \ref{tab:1}). The discriminant criteria we offer include accuracy, precision, recall, specificity, F1 score and G-mean. \\
(1)Accuracy=$\frac{\text{TP+TN}}{\text{TP+TN+FN+FP}}$\\
(2)Precision=$\frac{\text{TP}}{\text{TP+FP}}$\\
(3)Recall=$\frac{\text{TP}}{\text{TP+FN}}$\\
(4)Specificity=$\frac{\text{TN}}{\text{TN+FP}}$\\
% (5)Sensitivity=$\frac{\text{TP}}{\text{TP+FP}}$\\
(5)F1-score=$\frac{2 \cdot \text{Recall} \cdot \text{Precision}}{\text{Recall+Precision}}$\\
(6)G-mean=$\sqrt{\text{Recall}\cdot\text{Specificity}}$

\begin{table}[h]
\centering
\caption{Confusion matrix.}
\label{tab:1}
\setlength{\tabcolsep}{4mm}{
\begin{tabular}{l c c}
\toprule
$\quad$ &Predicted positive&Predicted negative\\
\midrule
True positive&TP&FN \\
True negative&FP&TN \\
\bottomrule
\end{tabular}}
\end{table}

\subsection{Experiments on Different Datasets}
In this section, to illustrate the universality of the proposed framework, we compare our proposed generalized framework, which combines standard first-order optimization algorithms SGD and second-order optimization algorithms oBFGS and oNAQ, with these algorithms on 12 different datasets. These datasets are diverse, including sparse, dense, imbalanced, high-dimensional, and low-dimensional. They were collected from the University of California, Irvine (UCI) Machine Learning repository\footnote{https://archive.ics.uci.edu}, Libsvm data\footnote{https://www.csie.ntu.edu.tw/\textasciitilde cjlin/libsvmtools/datasets/}. Details of the features are shown in Table \ref{tab:2}. The second column indicates the source of the dataset, the third column indicates the quantity of the dataset, the fourth column indicates the dimension, and the fifth column indicates the quantity of data categories. 

\begin{table}[h]
\centering
\caption{The description of the different datasets.}
\label{tab:2}
\setlength{\tabcolsep}{3.1mm}{
\begin{tabular}{l l l l l}
\toprule
Datasets&Source&Samples&Features&Classes\\
\midrule
A7a&UCI&16100&123&2\\
A8a&UCI&22696&123&2\\
A9a&UCI&32561&123&2\\
Mushroom&UCI&8124&22&2\\
Yeast&UCI&892&8&2\\
Ijcnn1&DP01a&49990&22&2\\
W1a&JP98a&2477&300&2\\
W2a&JP98a&3470&300&2\\
W3a&JP98a&4912&300&2\\
W4a&JP98a&7366&300&2\\
W5a&JP98a&9888&300&2\\
W6a&JP98a&17188&300&2\\
\bottomrule
\end{tabular}}
\end{table}

In this section, the parameters are set as follows. The learning rate of SGD optimization algorithm on different datasets $\epsilon$ is selected from \{0.1,0.3,0.5\}; The parameter $\tau$ related to the learning rate $\epsilon_{t}$ in oBFGS optimization algorithm is set to 10, $\epsilon_{t}=\frac{\tau}{\tau+t}$($t$ represents the $t$ iteration). In addition, to ensure the stability of the value, the change value of the independent variable is introduced in the calculation of the step difference, where the difference coefficient of the independent variable $\gamma$ is set to 0.2 on all datasets. Parameters $\tau$ related to the learning rate $\epsilon_{t}$ in the oNAQ optimization algorithm are set to 10, momentum coefficient $\mu$ to 0.1, and difference coefficient of independent variable $\gamma$ to 0.2 on all data sets. The highest limit of iterations and batch size of all algorithms are selected based on the size of the datasets. Refer to Table \ref{tab:3} for a comprehensive explanation.

\begin{figure}[h]
	\centering
	\subcaptionbox{A7a\label{fig6(a)}}{
		\includegraphics[width=5.2cm]{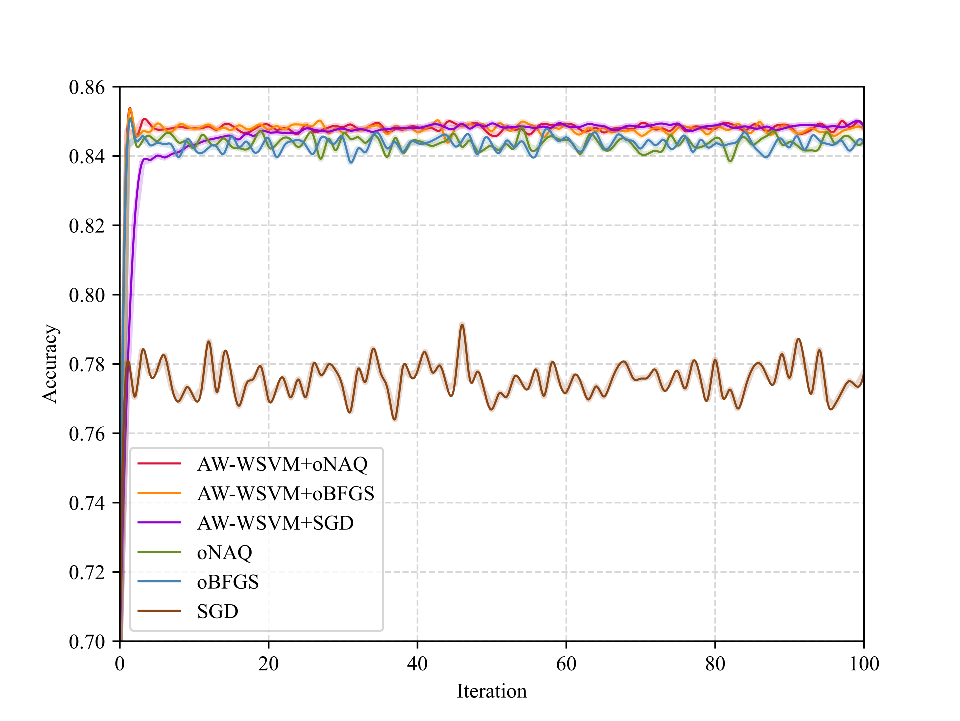}
	}
	\hfill 
	\subcaptionbox{A8a\label{fig6(b)}}{
		\includegraphics[width=5.2cm]{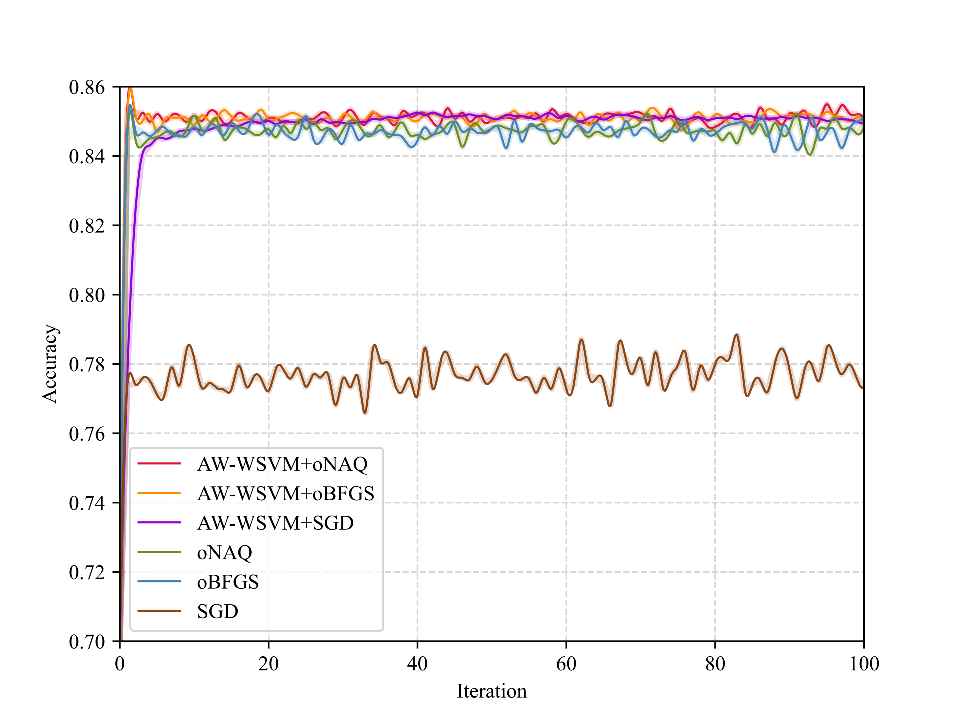}
	}
\hfill % 是为了让多幅图在一行均匀分布（不加的效果是都挤在中间）
	\subcaptionbox{A9a\label{fig6(c)}}{
		\includegraphics[width=5.2cm]{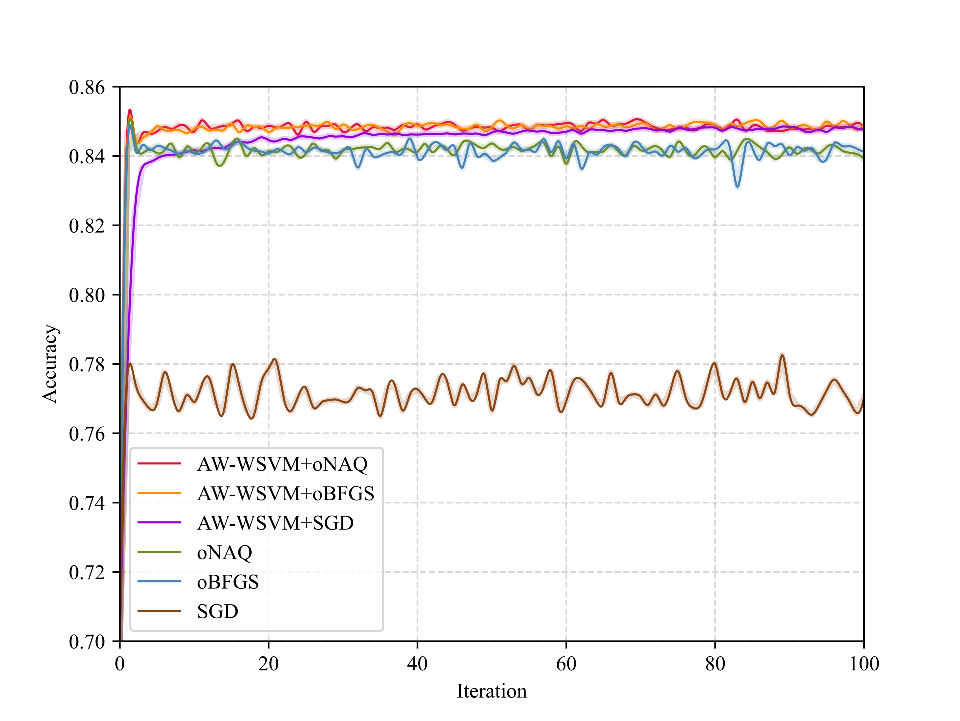}
	}
     \hfill 
	\subcaptionbox{Mushroom\label{fig6(d)}}{
		\includegraphics[width=5.2cm]{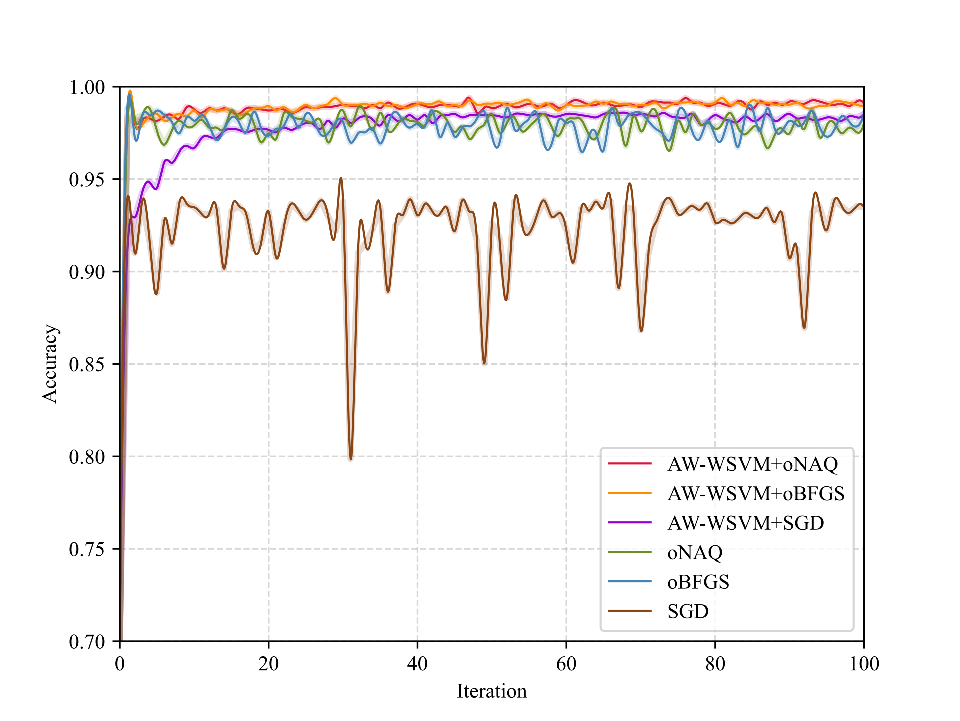}
	}
\hfill 
	\subcaptionbox{Yeast\label{fig6(e)}}{
		\includegraphics[width=5.2cm]{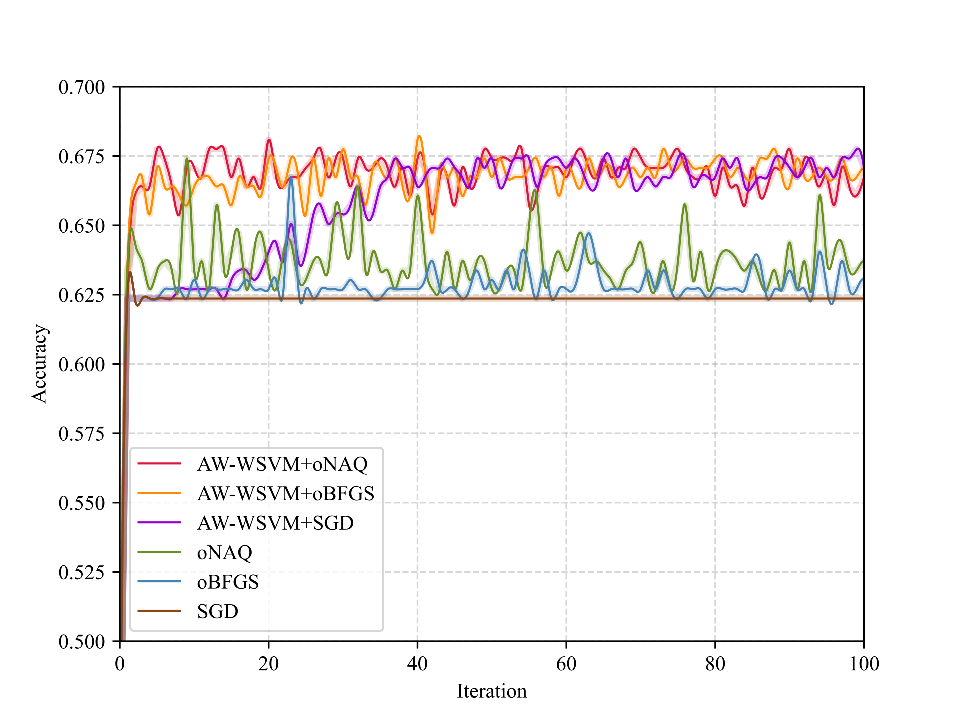}
	}
\hfill 
	\subcaptionbox{Ijcnn1\label{fig6(f)}}{
		\includegraphics[width=5.2cm]{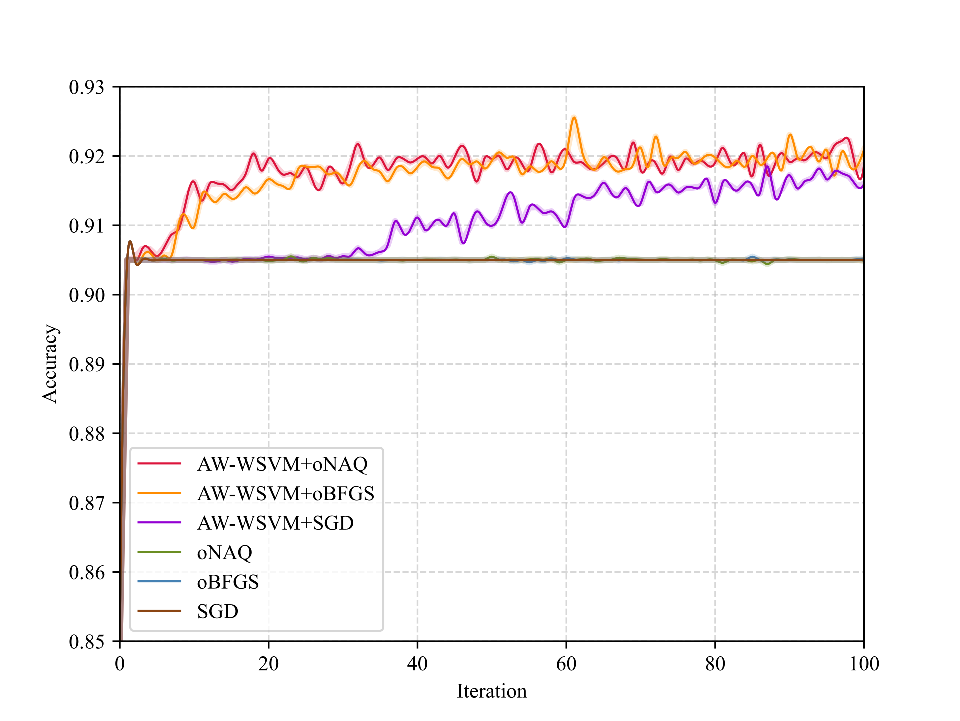}
	}
\hfill % 是为了让多幅图在一行均匀分布（不加的效果是都挤在中间）
	\subcaptionbox{W1a\label{fig6(g)}}{
		\includegraphics[width=5.2cm]{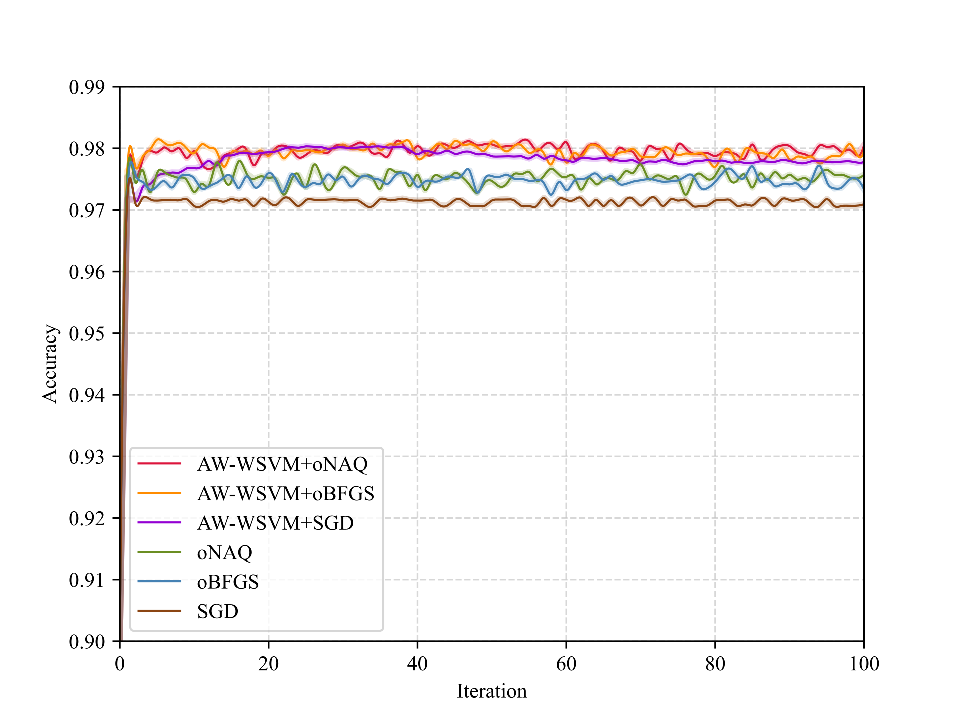}
	}
\hfill % 是为了让多幅图在一行均匀分布（不加的效果是都挤在中间）
	\subcaptionbox{W2a\label{fig6(h)}}{
		\includegraphics[width=5.2cm]{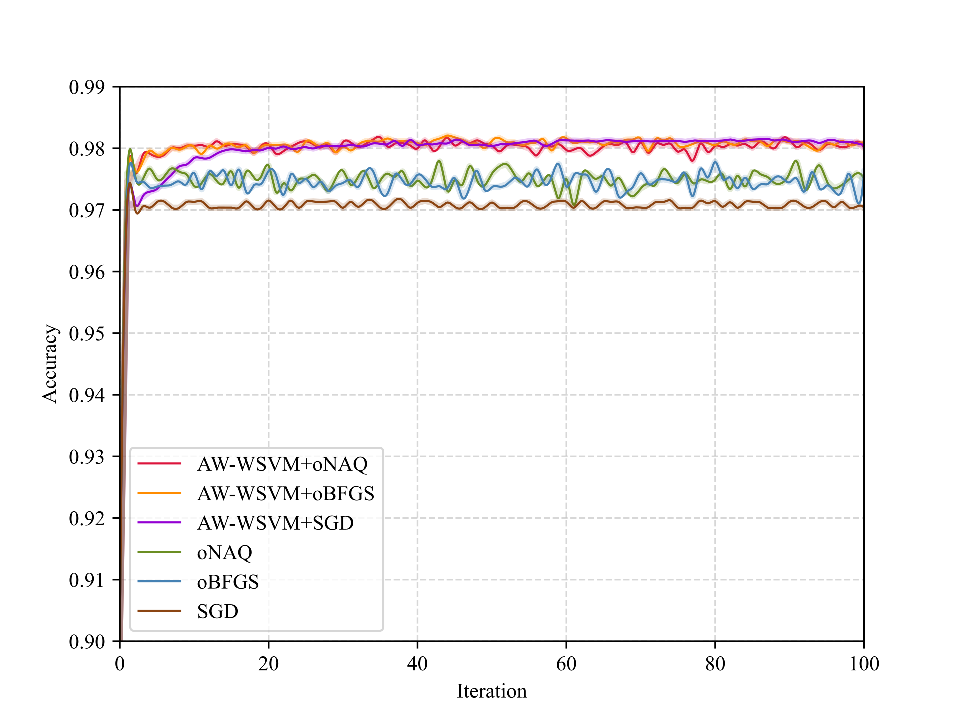}
	}
\hfill % 是为了让多幅图在一行均匀分布（不加的效果是都挤在中间）
	\subcaptionbox{W3a\label{fig6(i)}}{
		\includegraphics[width=5.2cm]{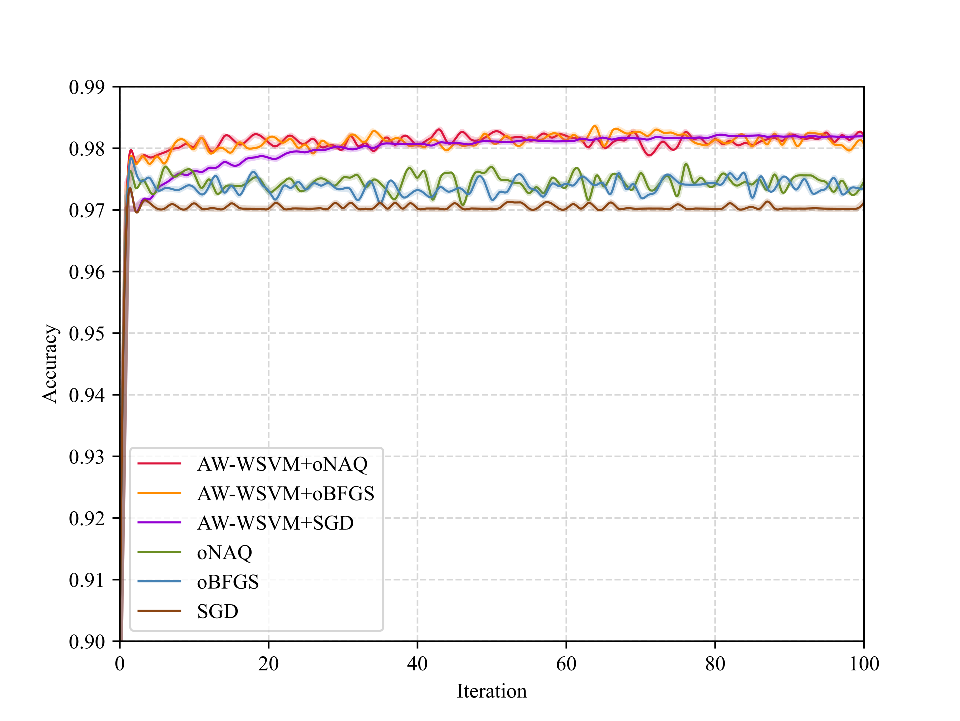}
	}
\hfill % 是为了让多幅图在一行均匀分布（不加的效果是都挤在中间）
	\subcaptionbox{W4a\label{fig6(j)}}{
		\includegraphics[width=5.2cm]{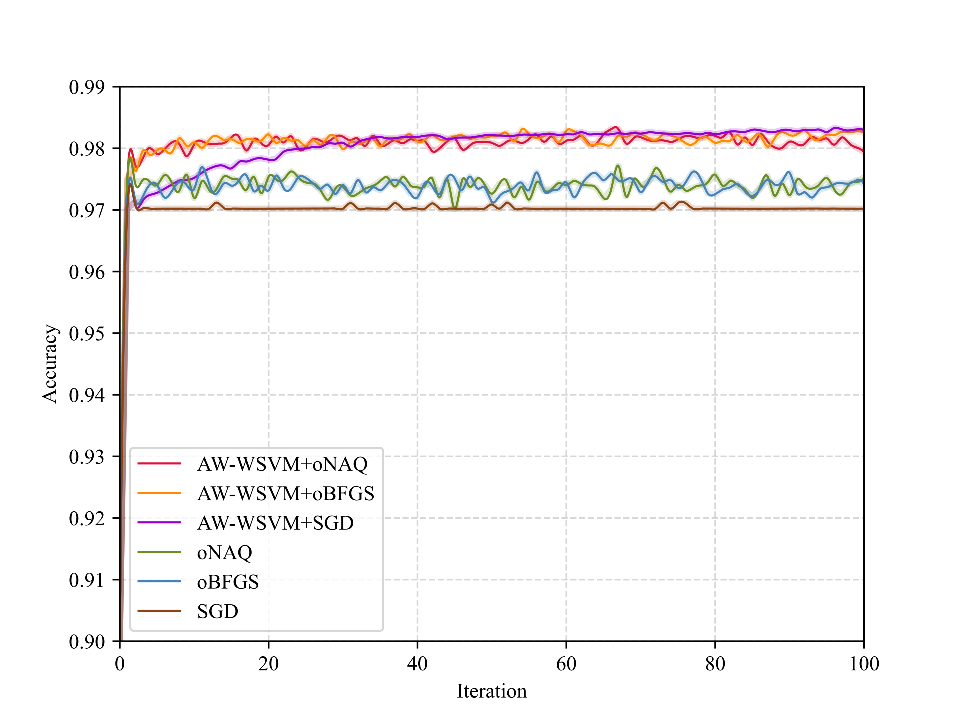}
	}
\hfill % 是为了让多幅图在一行均匀分布（不加的效果是都挤在中间）
	\subcaptionbox{W5a\label{fig6(k)}}{
		\includegraphics[width=5.2cm]{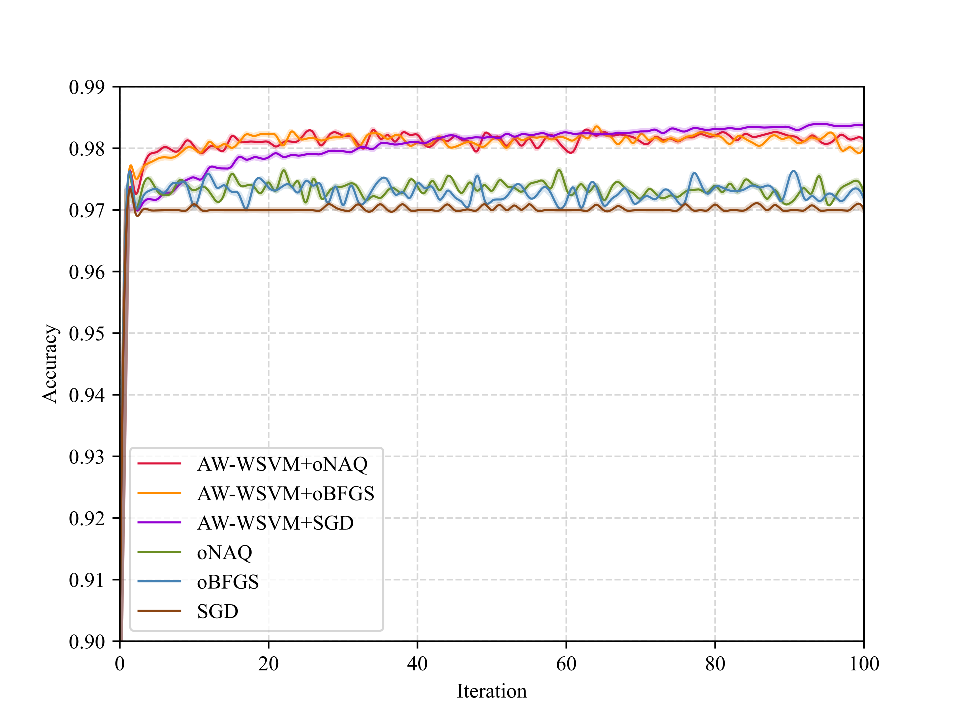}
	}
\hfill % 是为了让多幅图在一行均匀分布（不加的效果是都挤在中间）
	\subcaptionbox{W6a\label{fig6(l)}}{
		\includegraphics[width=5.2cm]{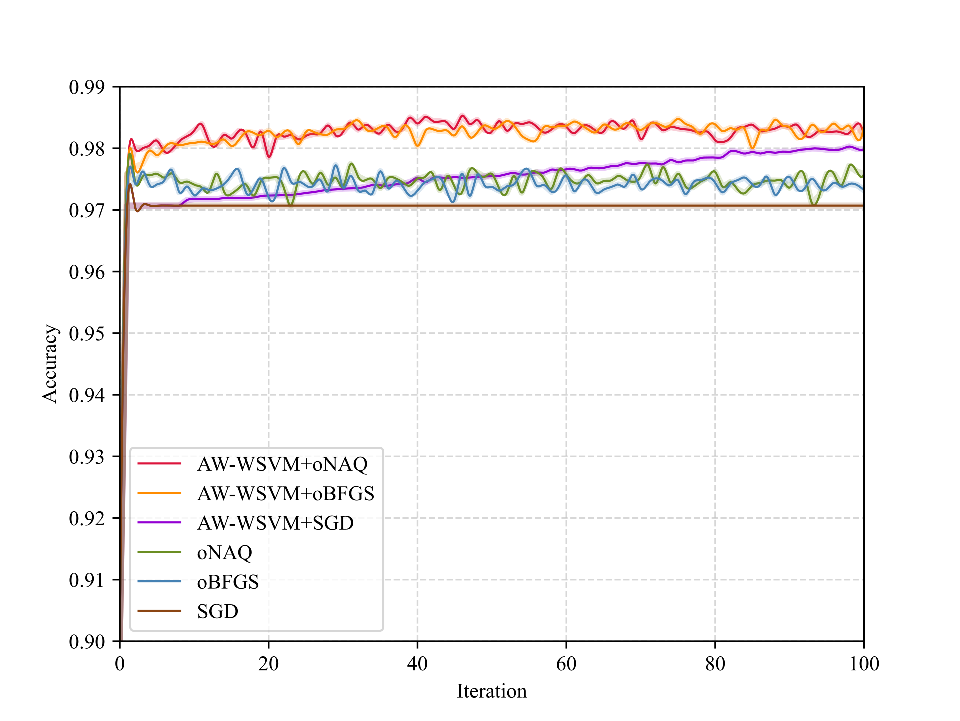}
	}
	\caption{Comparison of accuracy between standard algorithms and AW-WSVM on real datasets ($\sigma$=1).}
	\label{fig6}
\end{figure}

The variance $\sigma$ in the AW function is set to 1 empirically. We conducted 100 iterative experiments, taking the total amount of iterations as the horizontal coordinate and the test accuracy as the vertical coordinate to depict the performance of each algorithm, as shown in Fig.\ref{fig6}. Fig.\ref{fig6(a)}-\ref{fig6(l)} respectively plotted the experimental results of the 12 datasets in Table \ref{tab:2}. No matter whether the proposed method in the initial result performs well or not, it tends to be stable after a certain number of iterations. We can conclude that the proposed approach effectively improves the performance of SGD, oBFGS and oNAQ on all 12 datasets.

\begin{table}[h]
\centering
\caption{Details of parameter settings.}
\label{tab:3}
\setlength{\tabcolsep}{3mm}{
\begin{tabular}{l c c c c c c c c c c c c}
\toprule
Datasets&\multicolumn{3}{c}{SGD}&\multicolumn{4}{c}{oBFGS}&\multicolumn{5}{c}{oNAQ}\\
\cmidrule(r){2-4} \cmidrule(r){5-8} \cmidrule(r){9-13}
\quad&$\epsilon$&iter&Batchsize&$\tau$&$\gamma$&iter&Batchsize&$\tau$&$\mu$&$\gamma$&iter&Batchsize\\
\midrule
A7a&0.1&50&256&10&0.2&100&128&10&0.1&0.2&100&128\\
A8a&0.1&100&32&10&0.2&100&128&10&0.1&0.2&100&128\\
A9a&0.1&50&256&10&0.2&100&128&10&0.1&0.2&100&128\\
Mushroom&0.1&50&256&10&0.2&100&64&10&0.1&0.2&100&64\\
Yeast&0.3&50&128&10&0.2&50&64&10&0.1&0.2&50&64\\
Ijcnn1&0.5&50&64&10&0.2&50&64&10&0.1&0.2&50&64\\
W1a&0.5&50&64&10&0.2&50&64&10&0.1&0.2&50&64\\
W2a&0.5&50&64&10&0.2&50&64&10&0.1&0.2&50&64\\
W3a&0.5&50&64&10&0.2&50&64&10&0.1&0.2&50&64\\
W4a&0.5&50&64&10&0.2&50&64&10&0.1&0.2&50&64\\
W5a&0.5&50&64&10&0.2&50&64&10&0.1&0.2&50&64\\
W6a&0.5&50&64&10&0.2&50&64&10&0.1&0.2&50&64\\
\bottomrule
\end{tabular}}
\end{table}

We additionally evaluate the efficacy of the proposed framework, AW-WSVM, by using the confusion matrix. To ensure the stability of the results, we iterated 100 times. The metrics of different algorithms obtained from the confusion matrix can be presented in Table \ref{tab:4}-\ref{tab:7}, with the highest metrics bolded. Conclusions are easy to come by, and the proposed method outperforms SGD, oBFGS, and oNAQ on 12 datasets. The proposed method exhibits lower results on very few datasets. However, when considering the entirety of the recommended technique, it has successfully attained the intended outcome.

\begin{figure}[h]
\centering
	\subcaptionbox{A7a\label{gmzz1}}{
		\includegraphics[width=5.2cm]{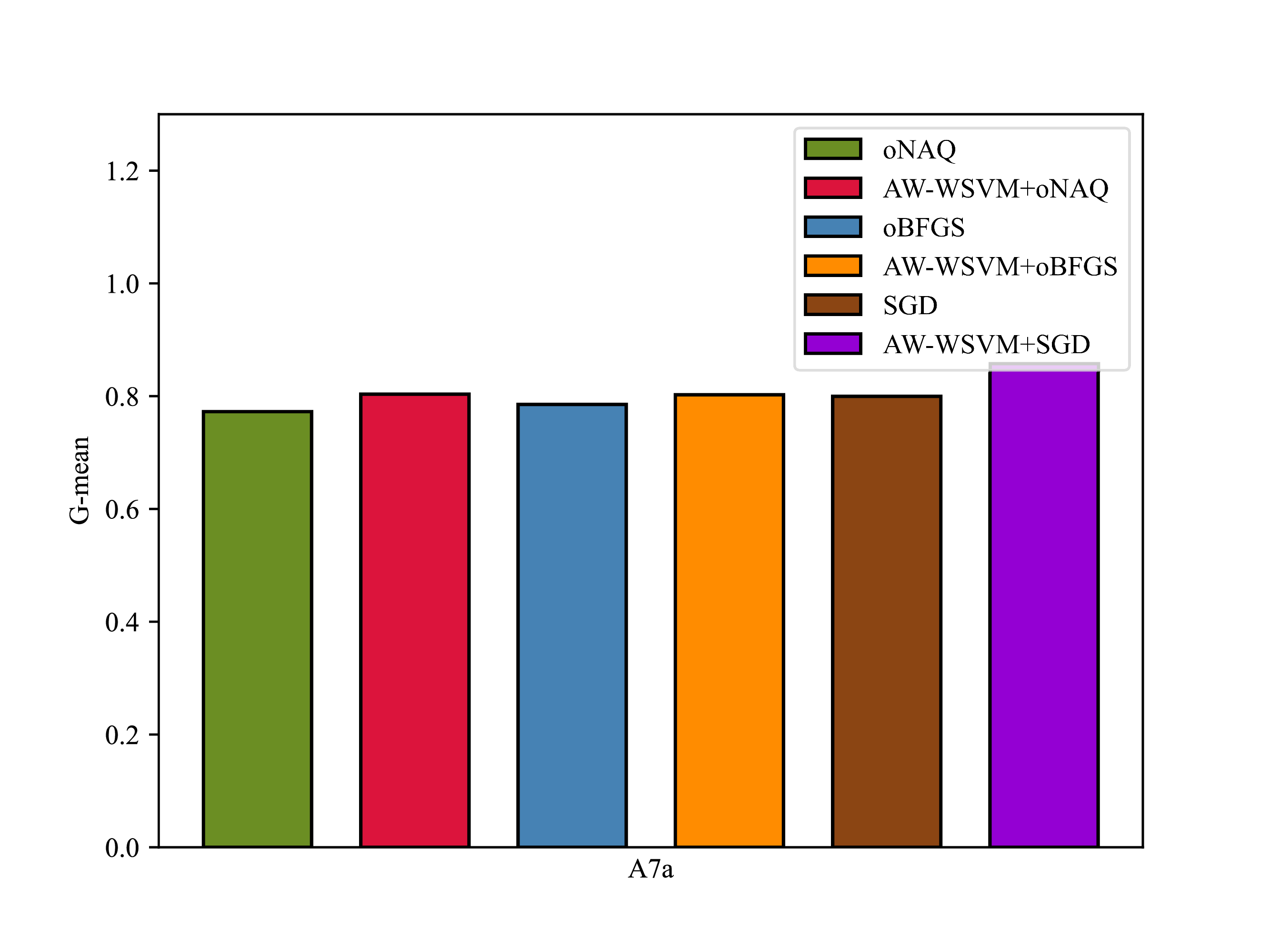}
	}
	\hfill 
    \subcaptionbox{A8a\label{gmzz2}}{
		\includegraphics[width=5.2cm]{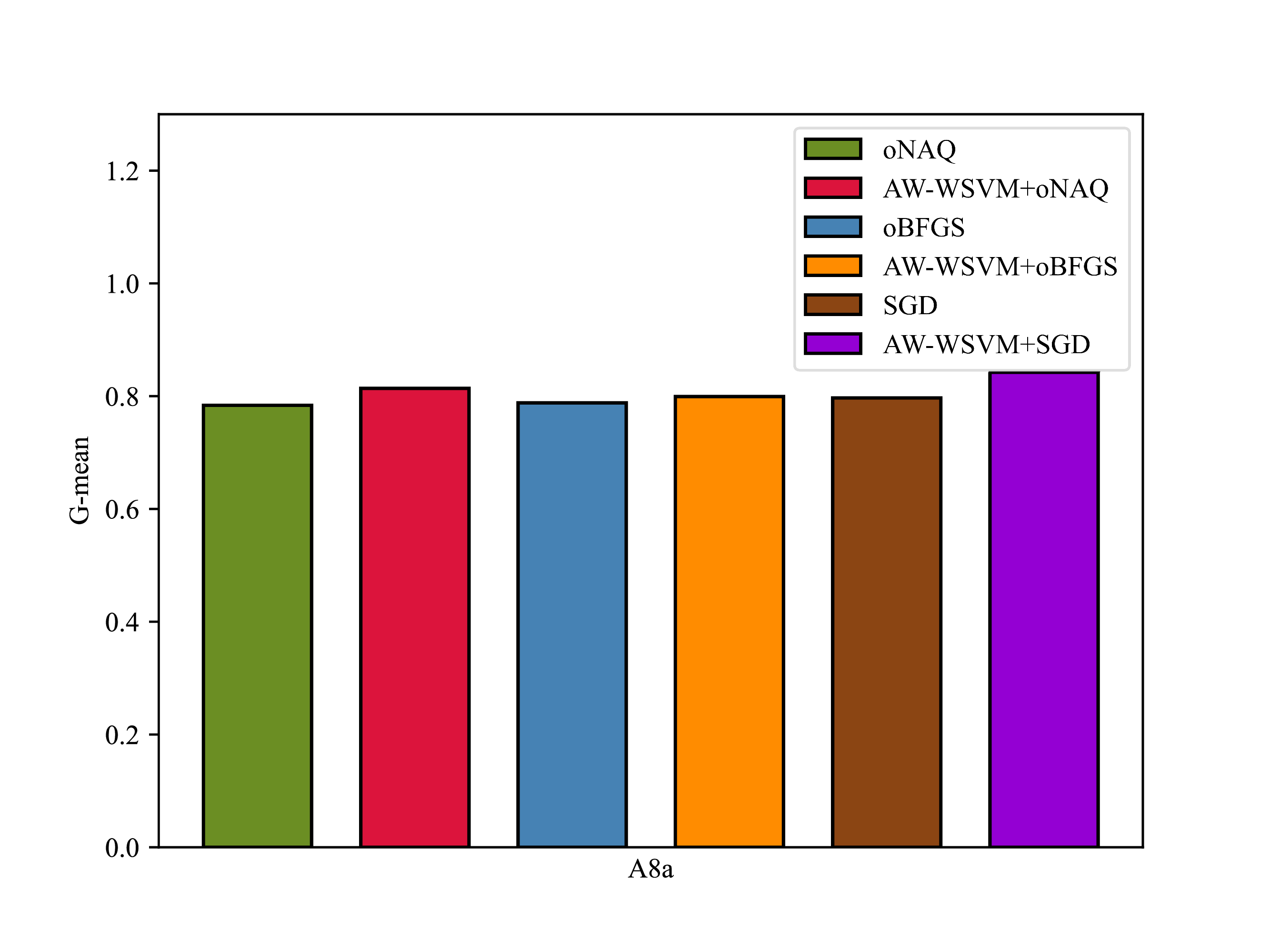}
	}
    \hfill 
    \subcaptionbox{A9a\label{gmzz3}}{
		\includegraphics[width=5.2cm]{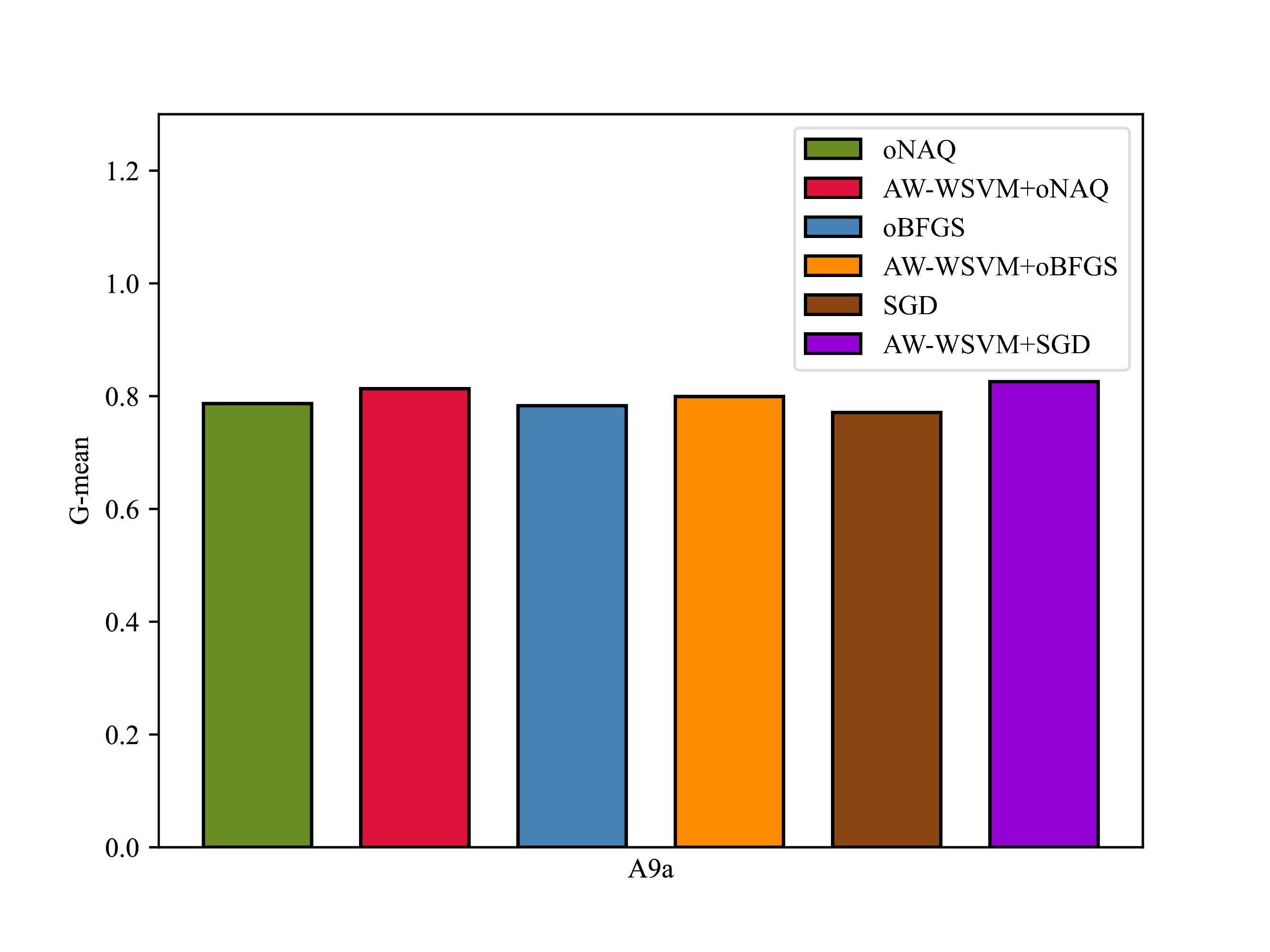}
	}
    \hfill 
    \subcaptionbox{Mushroom\label{gmzz4}}{
		\includegraphics[width=5.2cm]{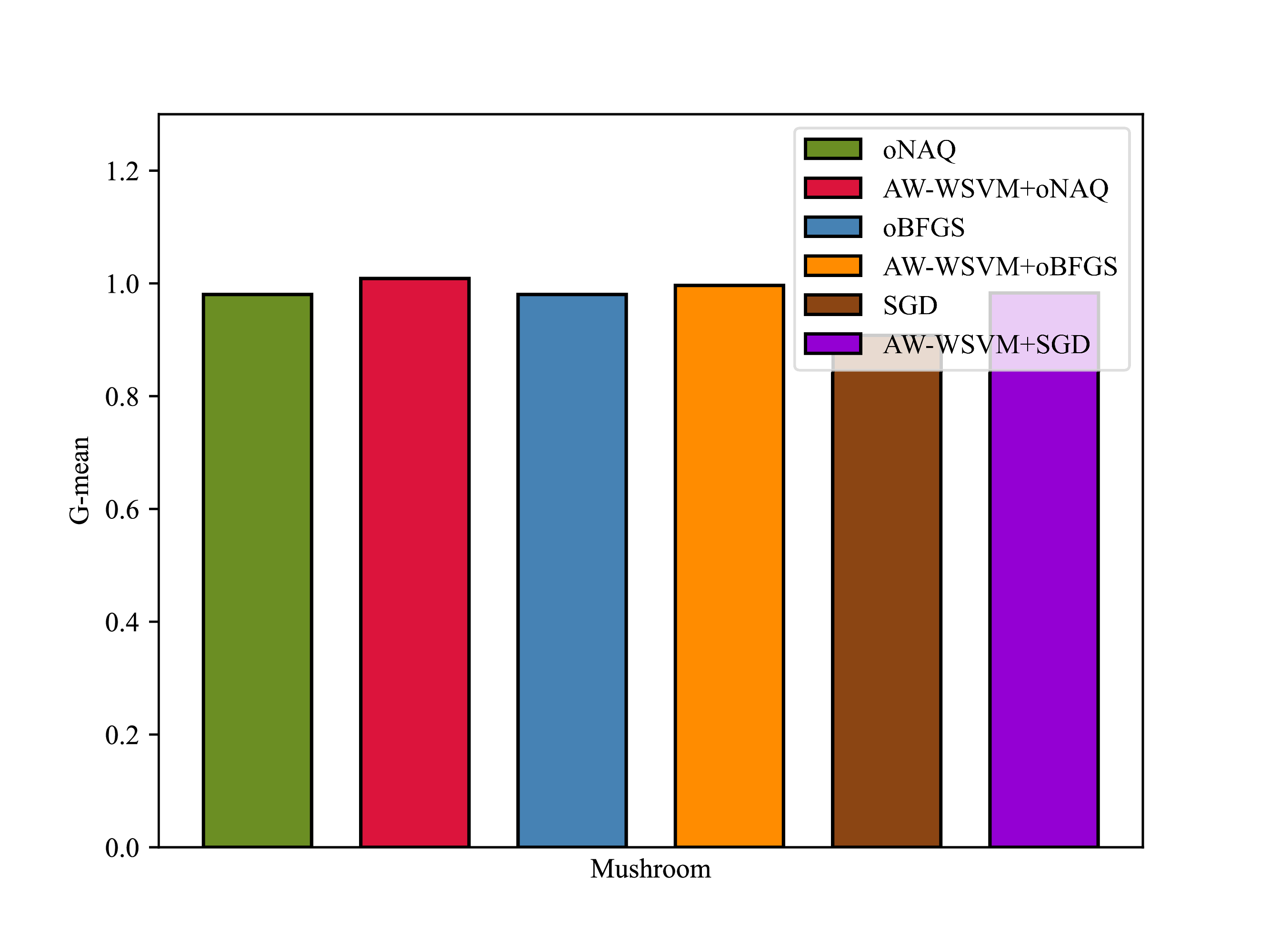}
	}
    \hfill 
    \subcaptionbox{Yeast\label{gmzz5}}{
		\includegraphics[width=5.2cm]{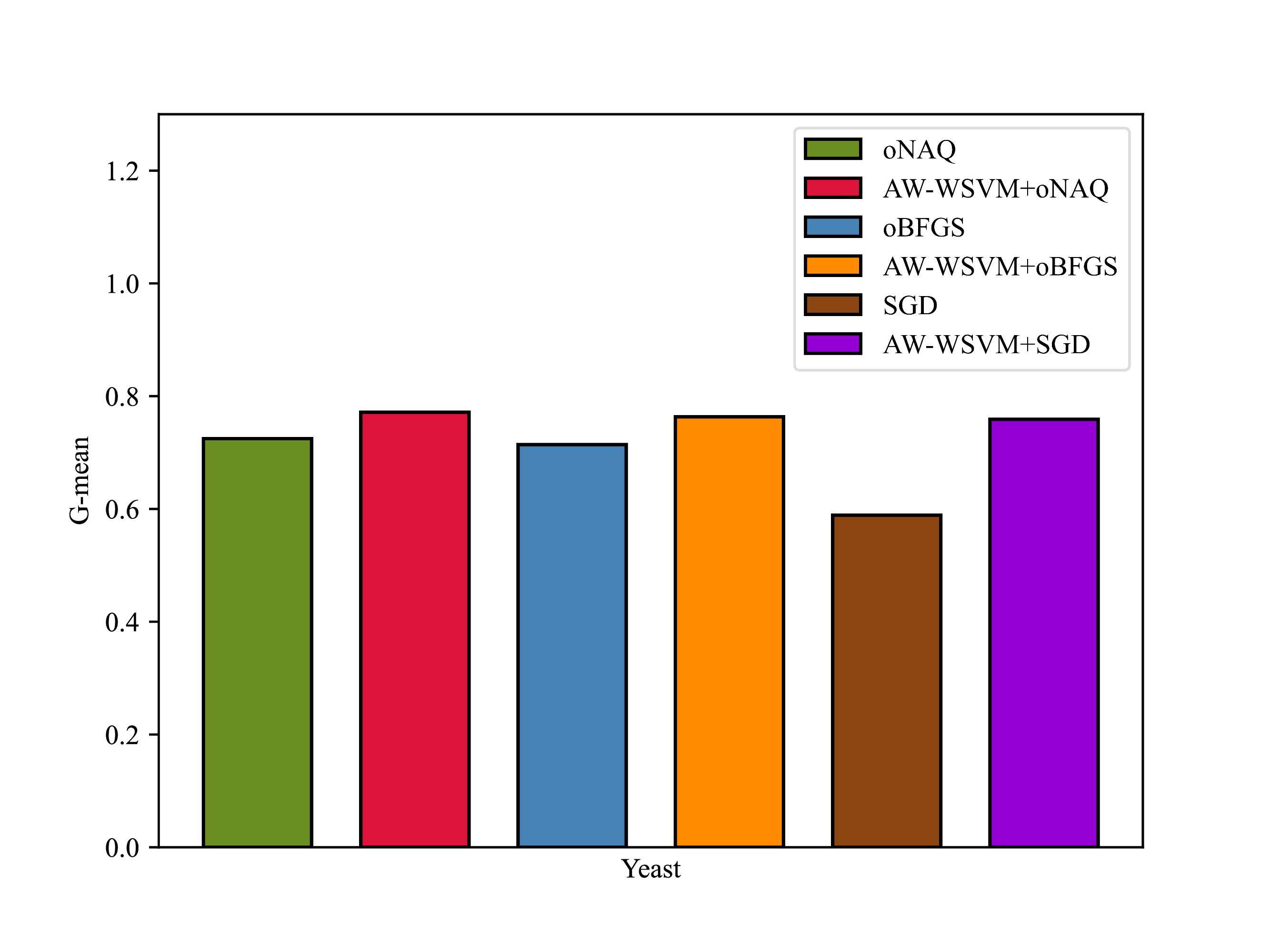}
	}
    \hfill 
    \subcaptionbox{Ijcnn1\label{gmzz6}}{
		\includegraphics[width=5.2cm]{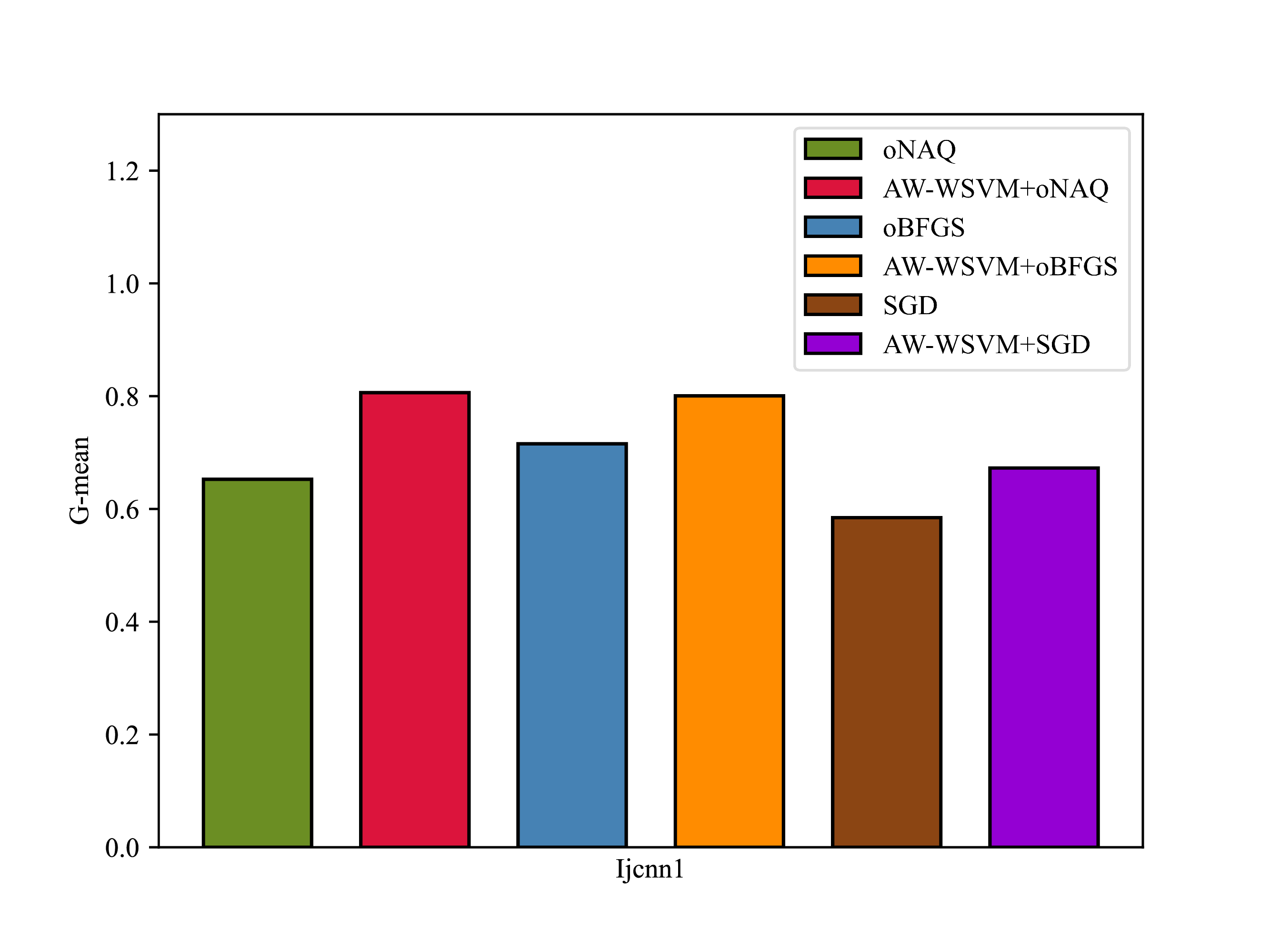}
	}
    \hfill 
    \subcaptionbox{W1a\label{gmzz7}}{
		\includegraphics[width=5.2cm]{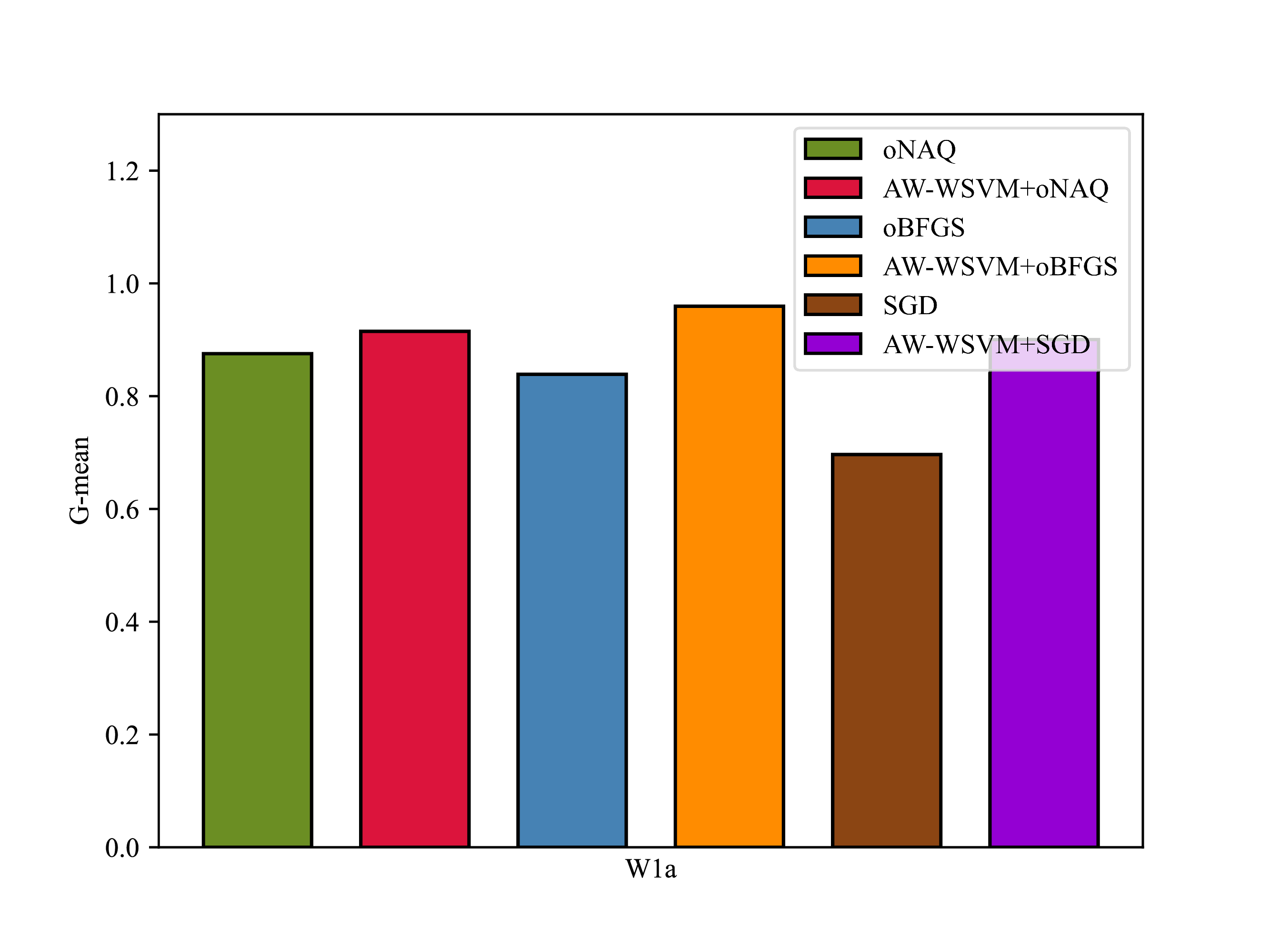}
	}
    \hfill 
    \subcaptionbox{W2a\label{gmzz8}}{
		\includegraphics[width=5.2cm]{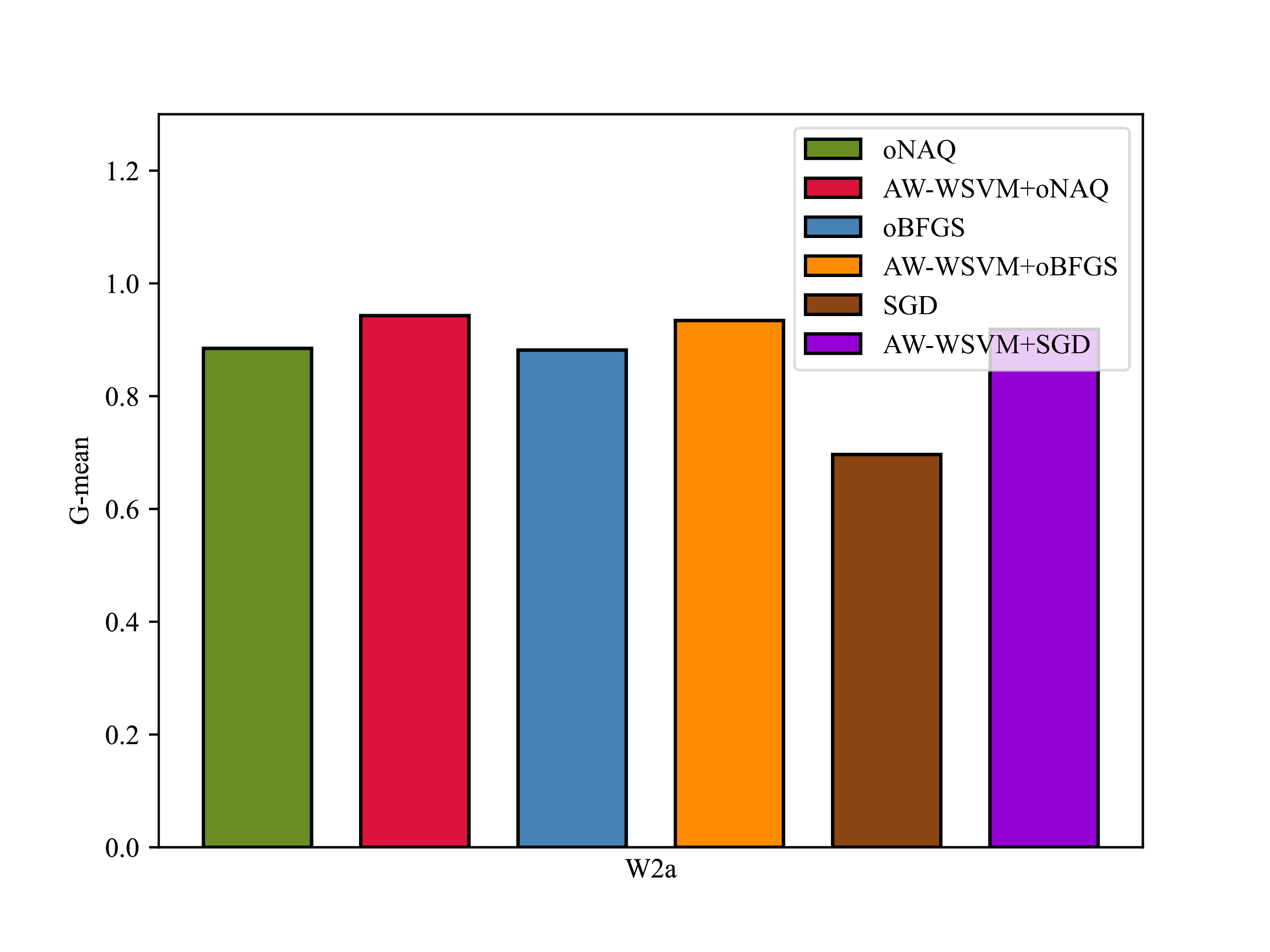}
	}
    \hfill 
    \subcaptionbox{W3a\label{gmzz9}}{
		\includegraphics[width=5.2cm]{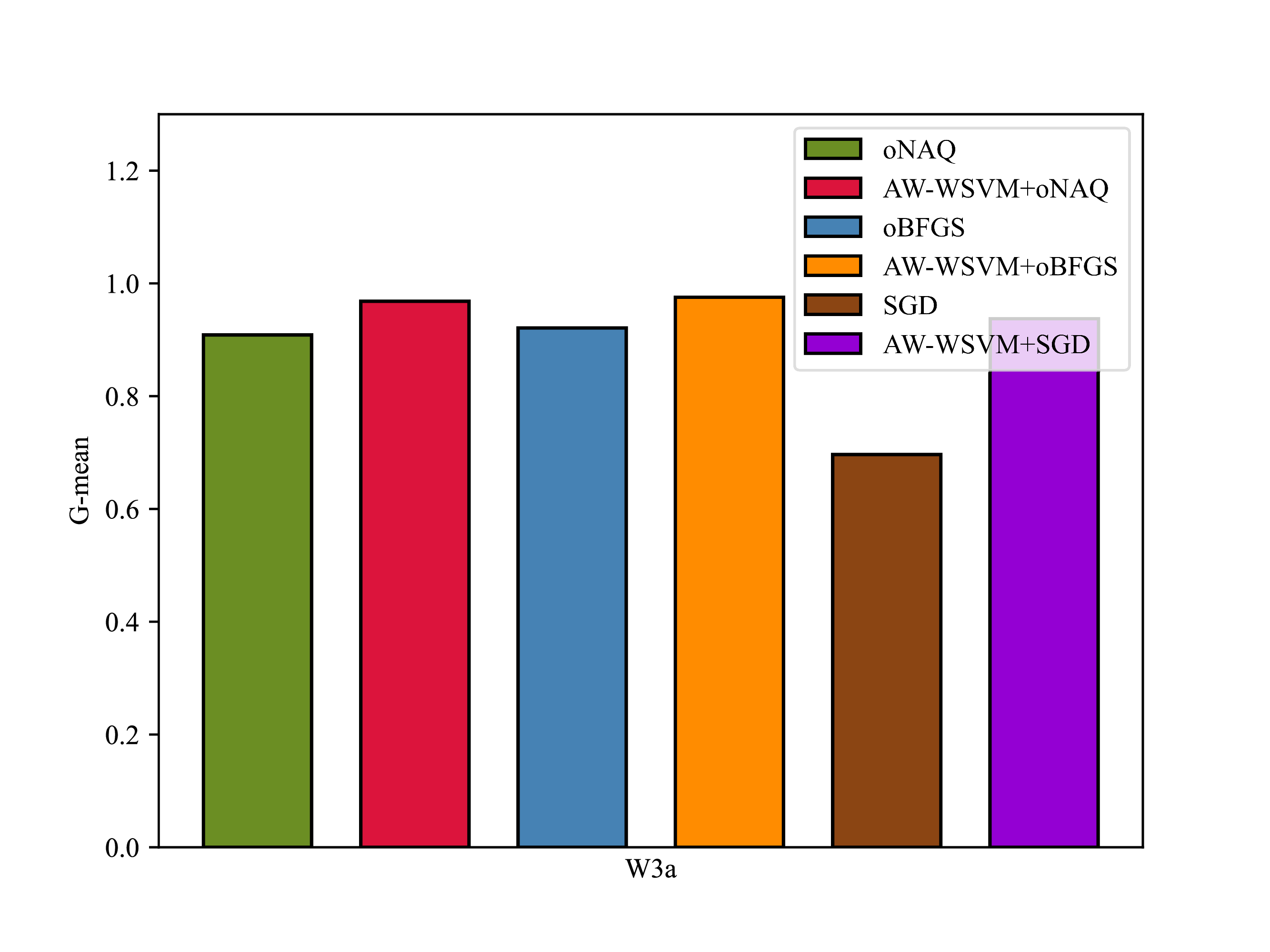}
	}
     \hfill 
    \subcaptionbox{W4a\label{gmzz10}}{
		\includegraphics[width=5.2cm]{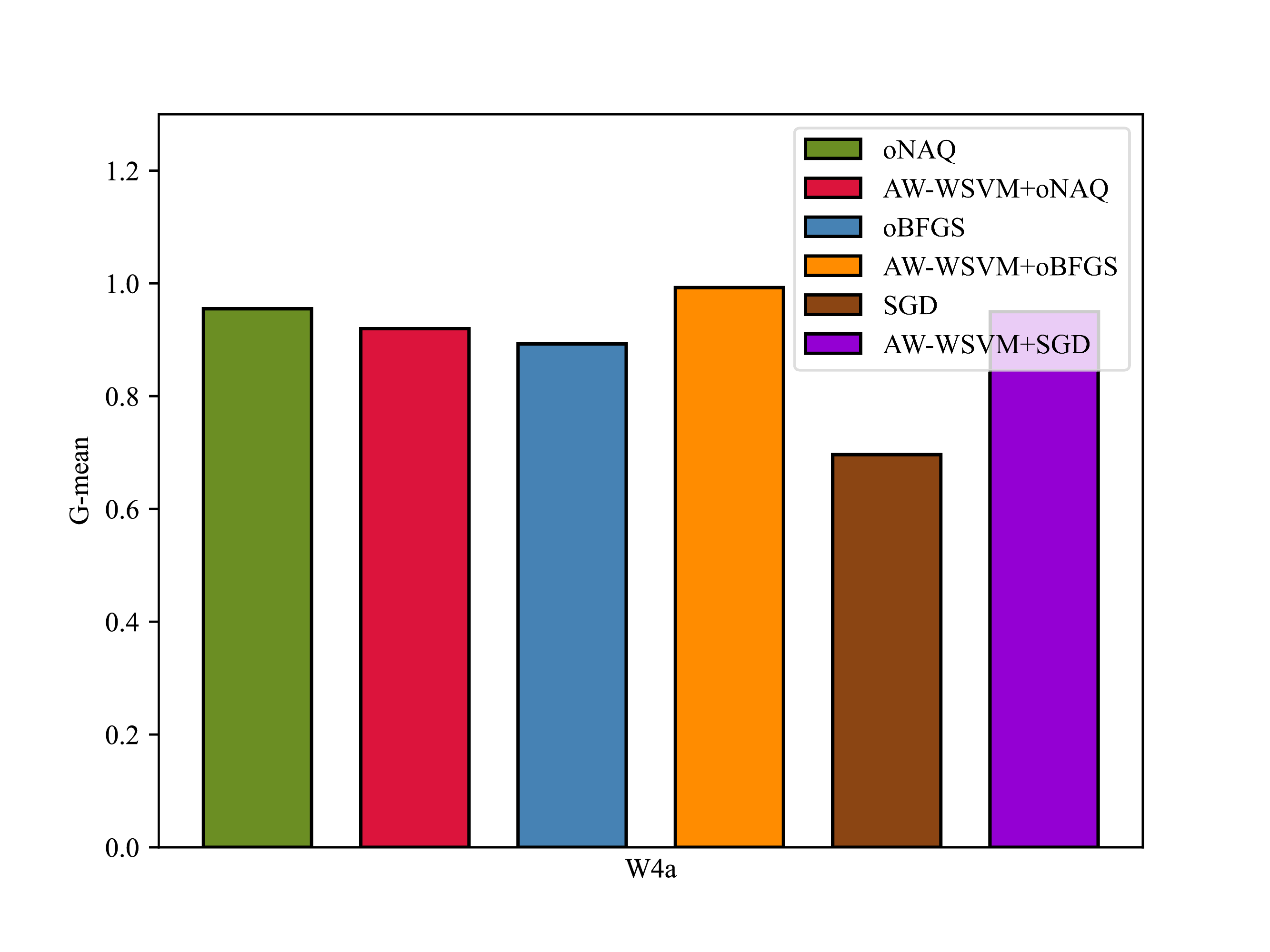}
	}
     \hfill 
    \subcaptionbox{W5a\label{gmzz11)}}{
		\includegraphics[width=5.2cm]{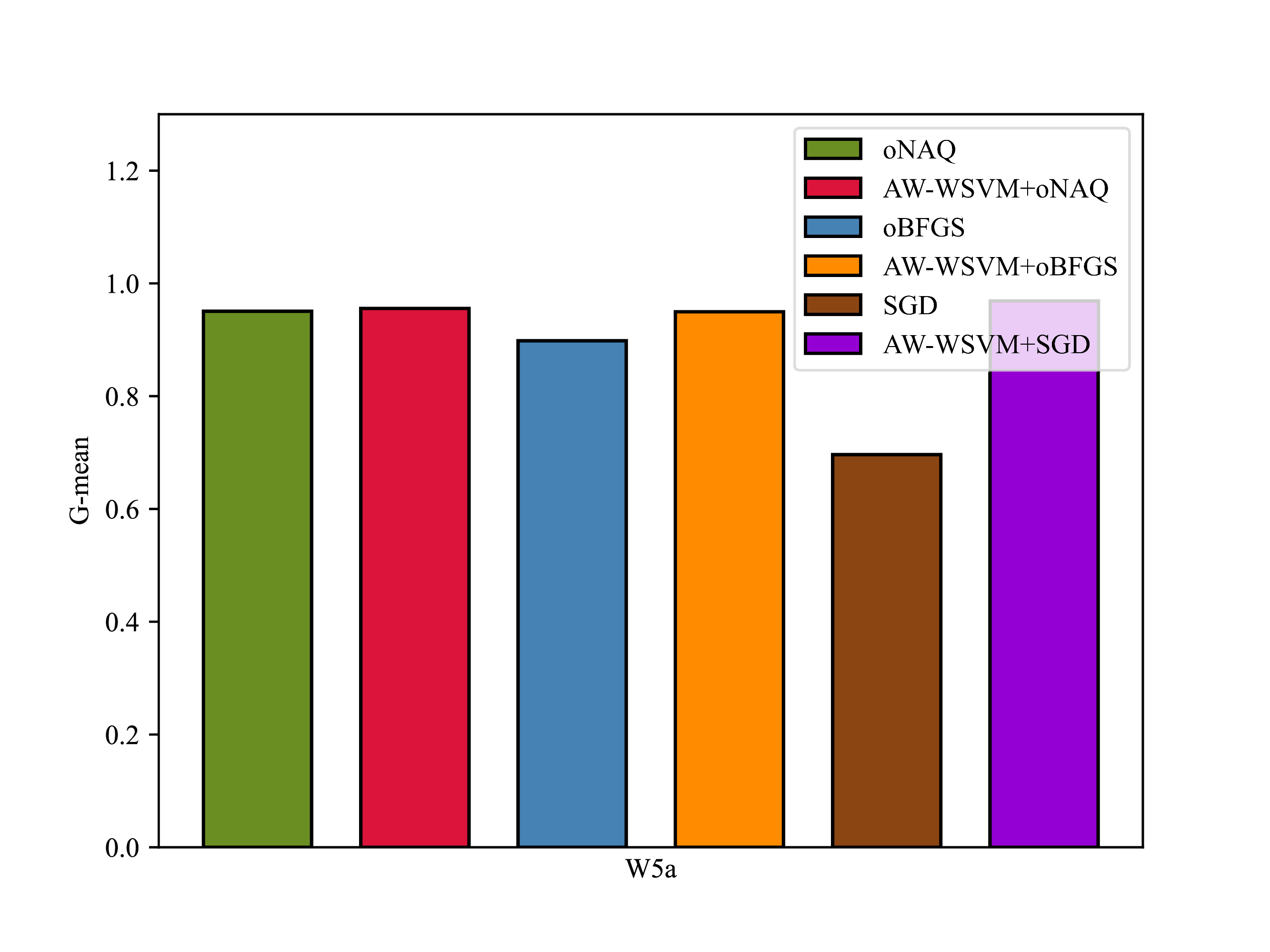}
	}
     \hfill 
    \subcaptionbox{W6a\label{gmzz12}}{
		\includegraphics[width=5.2cm]{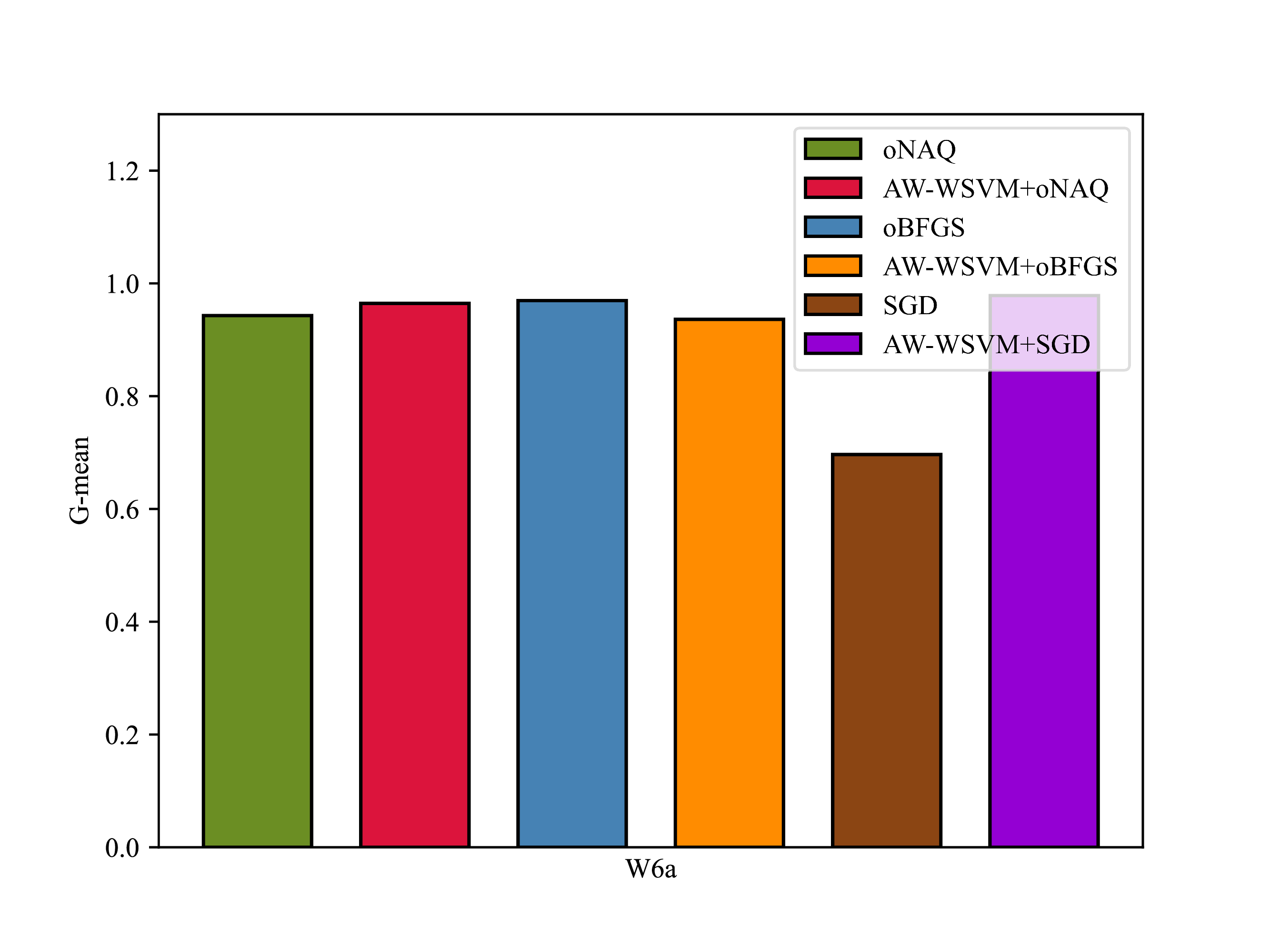}
	}
\caption{G-mean value.}
\label{gmzz} 
\end{figure}

In order to intuitively illustrate the effectiveness of our proposed method in processing these 12 datasets, we plotted a histogram of G-mean values. We randomly sampled 20 experimental results for each dataset, and the height of each bar in the histogram represents the expectation of G-mean value over these 20 sampled results. As shown in Fig.\ref{gmzz}, we can easily observe that our method performs better on all datasets.

\begin{table}[h]
\centering
\caption{Accuracy of six methods on different datasets.}
\label{tab:4}
\setlength{\tabcolsep}{3.5mm}{
\begin{tabular}{l c c c c c c}
\toprule
Datasets&oNAQ&AW-WSVM+oNAQ&oBFGS&AW-WSVM+oBFGS&SGD&AW-WSVM+SGD\\
\midrule
A7a&0.8431&\textbf{0.8486}&0.8453&\textbf{0.8469}&0.7682&\textbf{0.8486}\\
A8a&0.8492&\textbf{0.8514}&0.8456&\textbf{0.8521}&0.7767&\textbf{0.8513}\\
A9a&0.8425&\textbf{0.8445}&0.8434&\textbf{0.8469}&0.7733&\textbf{0.8500}\\
Mushroom&0.9832&\textbf{0.9906}&0.9791&\textbf{0.9865}&0.7733&\textbf{0.9869}\\
Yeast&0.6452&\textbf{0.6498}&0.6475&\textbf{0.6498}&\textbf{0.6495}&\textbf{0.6495}\\
Ijcnn1&0.9050&\textbf{0.9207}&0.9050&\textbf{0.9193}&0.9050&\textbf{0.9151}\\
W1a&0.9765&\textbf{0.9805}&0.9747&\textbf{0.9790}&0.9706&\textbf{0.9780}\\
W2a&0.9752&\textbf{0.9805}&0.9734&\textbf{0.9816}&0.9704&\textbf{0.9816}\\
W3a&0.9747&\textbf{0.9800}&0.9745&\textbf{0.9812}&0.9702&\textbf{0.9823}\\
W4a&0.9731&\textbf{0.9812}&0.9723&\textbf{0.9818}&0.9702&\textbf{0.9827}\\
W5a&0.9730&\textbf{0.9830}&0.9750&\textbf{0.9816}&0.9699&\textbf{0.9828}\\
W6a&0.9755&\textbf{0.9830}&0.9733&\textbf{0.9849}&0.9707&\textbf{0.9845}\\
\bottomrule
\end{tabular}}
\end{table}

\begin{table}[h]
\centering
\caption{G-mean of six methods on different datasets.}
\label{tab:5}
\setlength{\tabcolsep}{3.5mm}{
\begin{tabular}{l c c c c c c}
\toprule
Datasets&oNAQ&AW-WSVM+oNAQ&oBFGS&AW-WSVM+oBFGS&SGD&AW-WSVM+SGD\\
\midrule
A7a&0.7726&\textbf{0.7836}&0.7854&\textbf{0.7925}&0.7998&\textbf{0.8580}\\
A8a&0.7838&\textbf{0.7939}&0.7882&\textbf{0.7893}&0.7969&\textbf{0.8423}\\
A9a&0.7869&\textbf{0.7933}&0.7830&\textbf{0.7893}&0.7710&\textbf{0.8258}\\
Mushroom&0.9804&\textbf{0.9887}&0.9804&\textbf{0.9864}&0.9083&\textbf{0.9832}\\
Yeast&0.7247&\textbf{0.7515}&0.7140&\textbf{0.7534}&0.5889&\textbf{0.7590}\\
Ijcnn1&0.6526&\textbf{0.7863}&0.7155&\textbf{0.7906}&0.5846&\textbf{0.6726}\\
W1a&0.8754&\textbf{0.8949}&0.8388&\textbf{0.9495}&0.6966&\textbf{0.9009}\\
W2a&0.8848&\textbf{0.9228}&0.8819&\textbf{0.9241}&0.6965&\textbf{0.9189}\\
W3a&0.9086&\textbf{0.9483}&0.9210&\textbf{0.9654}&0.6965&\textbf{0.9375}\\
W4a&\textbf{0.9551}&0.8997&0.8926&\textbf{0.9825}&0.6964&\textbf{0.9501}\\
W5a&\textbf{0.9507}&0.9356&0.8983&\textbf{0.9397}&0.6963&\textbf{0.9693}\\
W6a&0.9429&\textbf{0.9445}&\textbf{0.9695}&0.9262&0.6966&\textbf{0.9787}\\
\bottomrule
\end{tabular}}
\end{table}

\begin{table}[h]
\centering
\caption{Recall of six methods on different datasets.}
\label{tab:6}
\setlength{\tabcolsep}{3.5mm}{
\begin{tabular}{l c c c c c c}
\toprule
Datasets&oNAQ&AW-WSVM+oNAQ&oBFGS&AW-WSVM+oBFGS&SGD&AW-WSVM+SGD\\
\midrule
A7a&0.8869&\textbf{0.8900}&0.8655&\textbf{0.8846}&0.7713&\textbf{0.8769}\\
A8a&0.8725&\textbf{0.8800}&0.8840&\textbf{0.8859}&0.7698&\textbf{0.8846}\\
A9a&0.8735&\textbf{0.8825}&0.8772&\textbf{0.8814}&0.7728&\textbf{0.8745}\\
Mushroom&0.9862&\textbf{0.9908}&\textbf{0.9899}&0.9863&0.8654&\textbf{0.9832}\\
Yeast&0.7333&\textbf{0.7727}&0.7142&\textbf{0.7692}&0.5000&\textbf{0.7777}\\
Ijcnn1&0.9056&\textbf{0.9297}&0.9058&\textbf{0.9332}&0.9049&\textbf{0.9052}\\
W1a&0.9769&\textbf{0.9841}&0.9734&\textbf{0.9851}&0.9706&\textbf{0.9820}\\
W2a&0.9756&\textbf{0.9831}&0.9769&\textbf{0.9844}&0.9703&\textbf{0.9815}\\
W3a&0.9748&\textbf{0.9844}&0.9736&\textbf{0.9823}&0.9702&\textbf{0.9793}\\
W4a&0.9755&\textbf{0.9815}&0.9752&\textbf{0.9867}&0.9702&\textbf{0.9796}\\
W5a&0.9728&\textbf{0.9826}&0.8983&\textbf{0.9373}&0.9699&\textbf{0.9787}\\
W6a&0.9762&\textbf{0.9831}&0.9732&\textbf{0.9857}&0.9707&\textbf{0.9798}\\
\bottomrule
\end{tabular}}
\end{table}

\begin{table}[h]
\centering
\caption{F1-score of six methods on different datasets.}
\label{tab:7}
\setlength{\tabcolsep}{3.5mm}{
\begin{tabular}{l c c c c c c}
\toprule
Datasets&oNAQ&AW-WSVM+oNAQ&oBFGS&AW-WSVM+oBFGS&SGD&AW-WSVM+SGD\\
\midrule
A7a&0.8955&\textbf{0.8996}&0.8976&\textbf{0.9025}&0.8706&\textbf{0.9038}\\
A8a&0.8987&\textbf{0.9030}&0.9016&\textbf{0.9022}&0.8698&\textbf{0.9046}\\
A9a&0.9000&\textbf{0.9033}&0.8992&\textbf{0.9018}&0.8716&\textbf{0.9048}\\
Mushroom&0.9822&\textbf{0.9897}&0.9826&\textbf{0.9874}&0.9129&\textbf{0.9850}\\
Yeast&0.6241&\textbf{0.6285}&0.6277&\textbf{0.6293}&\textbf{0.6004}&\textbf{0.6004}\\
Ijcnn1&0.9500&\textbf{0.9563}&0.9502&\textbf{0.9574}&\textbf{0.9501}&0.9499\\
W1a&0.9876&\textbf{0.9900}&0.9864&\textbf{0.9895}&0.9706&\textbf{0.9820}\\
W2a&0.9872&\textbf{0.9898}&0.9878&\textbf{0.9902}&0.9849&\textbf{0.9897}\\
W3a&0.9870&\textbf{0.9907}&0.9865&\textbf{0.9901}&0.9848&\textbf{0.9890}\\
W4a&0.9874&\textbf{0.9894}&0.9868&\textbf{0.9910}&0.9848&\textbf{0.9893}\\
W5a&0.9861&\textbf{0.9904}&0.9856&\textbf{0.9899}&0.9847&\textbf{0.9890}\\
W6a&0.9876&\textbf{0.9908}&0.9863&\textbf{0.9916}&0.9851&\textbf{0.9897}\\
\bottomrule
\end{tabular}}
\end{table}

\subsection{Statistical Comparison}

This section makes a statistical comparison \cite{36} of the experimental results. The significance of performance differences between classifiers, including test accuracy and prediction accuracy, were tested using Friedman tests and Nemenyi post hoc tests \cite{34,35}. The Friedman test is utilized to compare the mean ranking of several algorithms on N datasets. Define $r_{i}^{j}$ as the ranking of the $j$ th algorithm on the $i$ th dataset out of a total of N datasets, and $R_{j}=\frac{1}{N}\sum_{i=1}^{N}r_{i}^{j}$. According to the null hypothesis, all algorithms are considered to be identical, so their average ranking $R_{j}$ should be the same. The Friedman statistic $R_{j}$ follows a Chi-square distribution with K-1 degrees of freedom

\begin{equation}
    \chi_{F}^{2}=\frac{12N}{K(K+1)}[R_{j}^{2}-\frac{(K(K+1)^{2})}{4}]
\end{equation}
where $K$ is the number of algorithms involved in the comparison.
\par Table \ref{tab:8} shows the average ranking $R_{j}$ of the six classifiers in test accuracy. The p-value of Friedman's test is $1.7122\times10^{-9}$. The P-value is less than the threshold for significance level $\alpha=0.05$, indicating that the null assumption should be discarded. The above results show that there are great differences among all classification methods from a statistical point of view. We compared the g-mean of the six classifiers (see Table \ref{tab:9}), with a p-value of $1.1560\times10^{-5}$ for Friedman's test, indicating significant differences among all classification methods. Table \ref{tab:10}-\ref{tab:11} record the Recall and F1-score of six classifiers. The Friedman's test p-values for Recall and F1 are $1.4607\times10^{-7}$ and $2.4246\times10^{-8}$, respectively, which indicates significant differences between the methods.

Consequently, to determine the classifiers that exhibit substantial differences, it is essential to perform a post-hoc test in order to further differentiate between the methods. Nemenyi post hoc test is selected in this paper. The Nemenyi test calculates the critical range of the difference in average order values
\begin{equation}
   CD=q_{\alpha}\sqrt{\frac{K(K+1)}{6N}} 
\end{equation}

If the discrepancy among the mean rank measurements of the two distinct algorithms exceeds the critical range (CD), the hypothesis that the two algorithms exhibit comparable results is refuted with the equivalent level of confidence. The variables $K$ and $N$ correspond to the quantities of algorithms and datasets, correspondingly. When the significant level $\alpha$=0.05, $q_{\alpha}$=2.850. So we get CD = 2.85×$\sqrt{6(6+1)/(6\times12)}$=2.1767.

\begin{table}[h]
\centering
\caption{Friedman ranking of these methods according to prediction accuracies on different datasets.}
\label{tab:8}
\setlength{\tabcolsep}{3.5mm}{
\begin{tabular}{l c c c c c c}
\toprule
Datasets&oNAQ&AW-WSVM+oNAQ&oBFGS&AW-WSVM+oBFGS&SGD&AW-WSVM+SGD\\
\midrule
A7a&4&1.5&3&2&5&1.5\\
A8a&4&2&5&1&6&3\\
A9a&4&2&5&1&6&3\\
Mushroom&4&1&5&3&6&2\\
Yeast&4&1.5&3&1.5&2.5&2.5\\
Ijcnn1&4.5&1&4.5&2&4.5&3\\
W1a&4&1&5&2&6&3\\
W2a&3&2&4&1.5&5&1.5\\
W3a&4&3&5&2&6&1\\
W4a&4&3&5&2&6&1\\
W5a&5&1&4&3&6&2\\
W6a&4&3&5&1&6&2\\
\hline
R&4.04&1.83&4.46&1.83&5.42&2.13\\
\bottomrule
\end{tabular}}
\end{table}

\begin{figure}[h]
	\subcaptionbox{Classification accuracy\label{fig7(a)}}{
		\includegraphics[width=8cm]{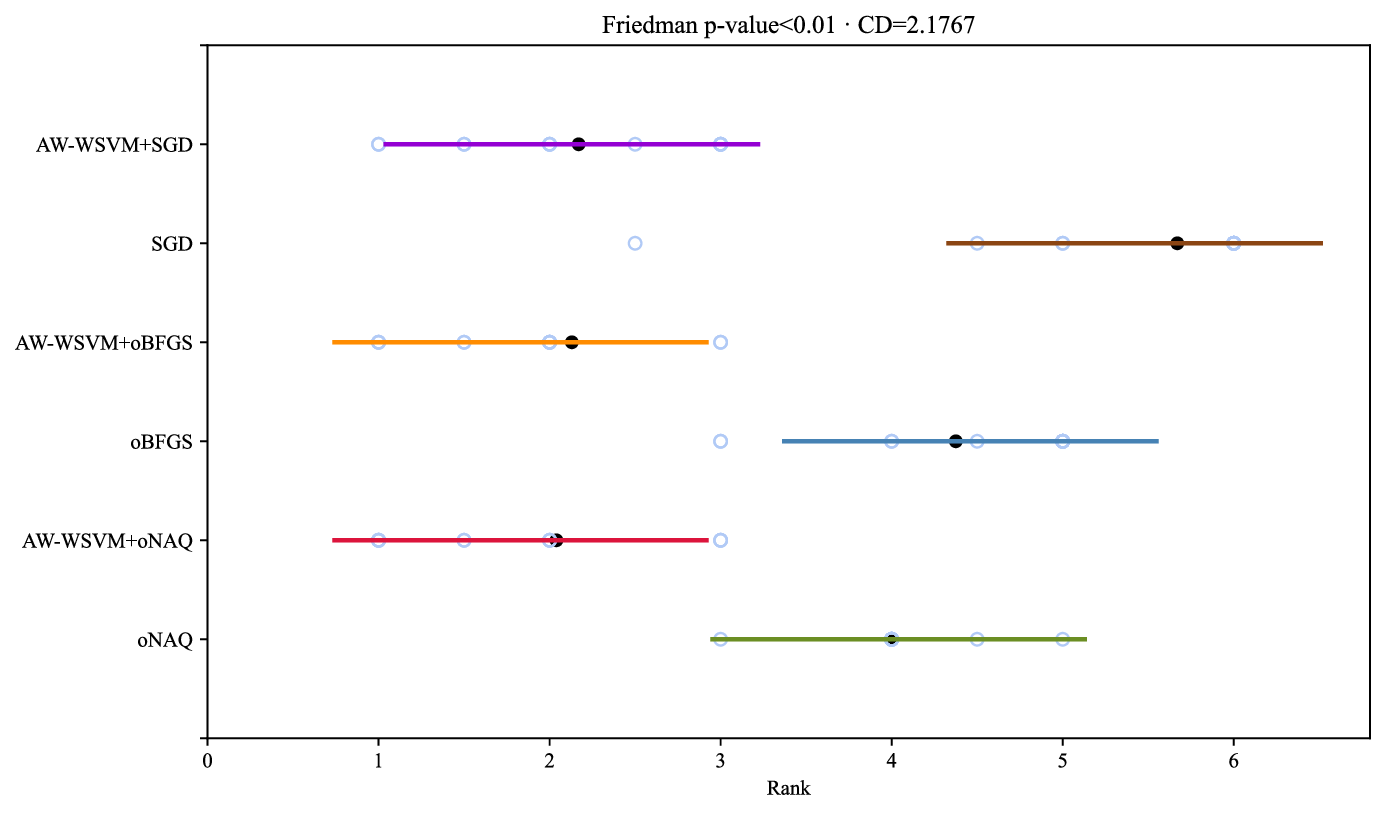}
	}
	\hfill % 是为了让多幅图在一行均匀分布（不加的效果是都挤在中间）
	\subcaptionbox{Classification G-mean\label{fig7(b)}}{
		\includegraphics[width=8cm]{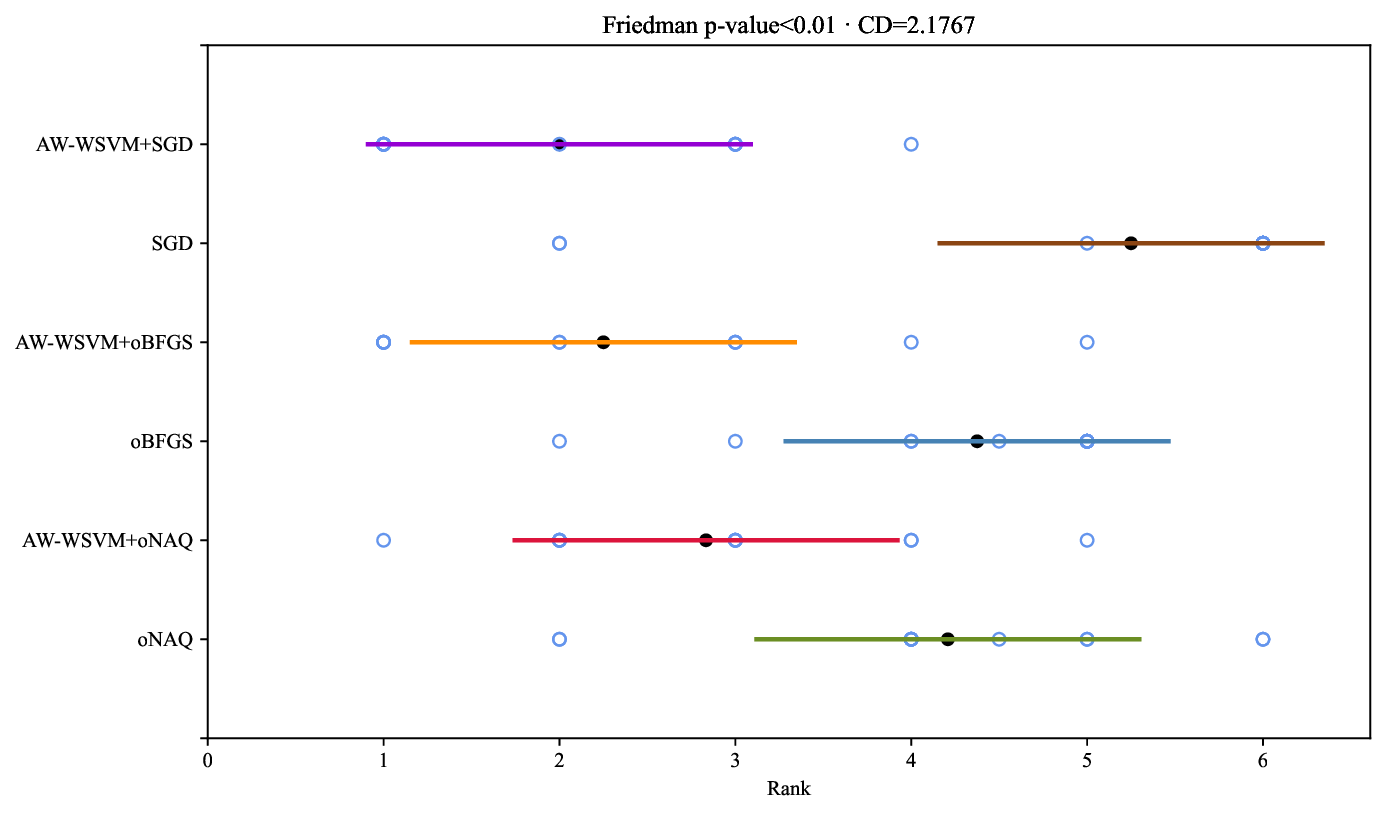}
	}
    \hfill % 是为了让多幅图在一行均匀分布（不加的效果是都挤在中间）
	\subcaptionbox{Classification Recall\label{fig7(c)}}{
		\includegraphics[width=8cm]{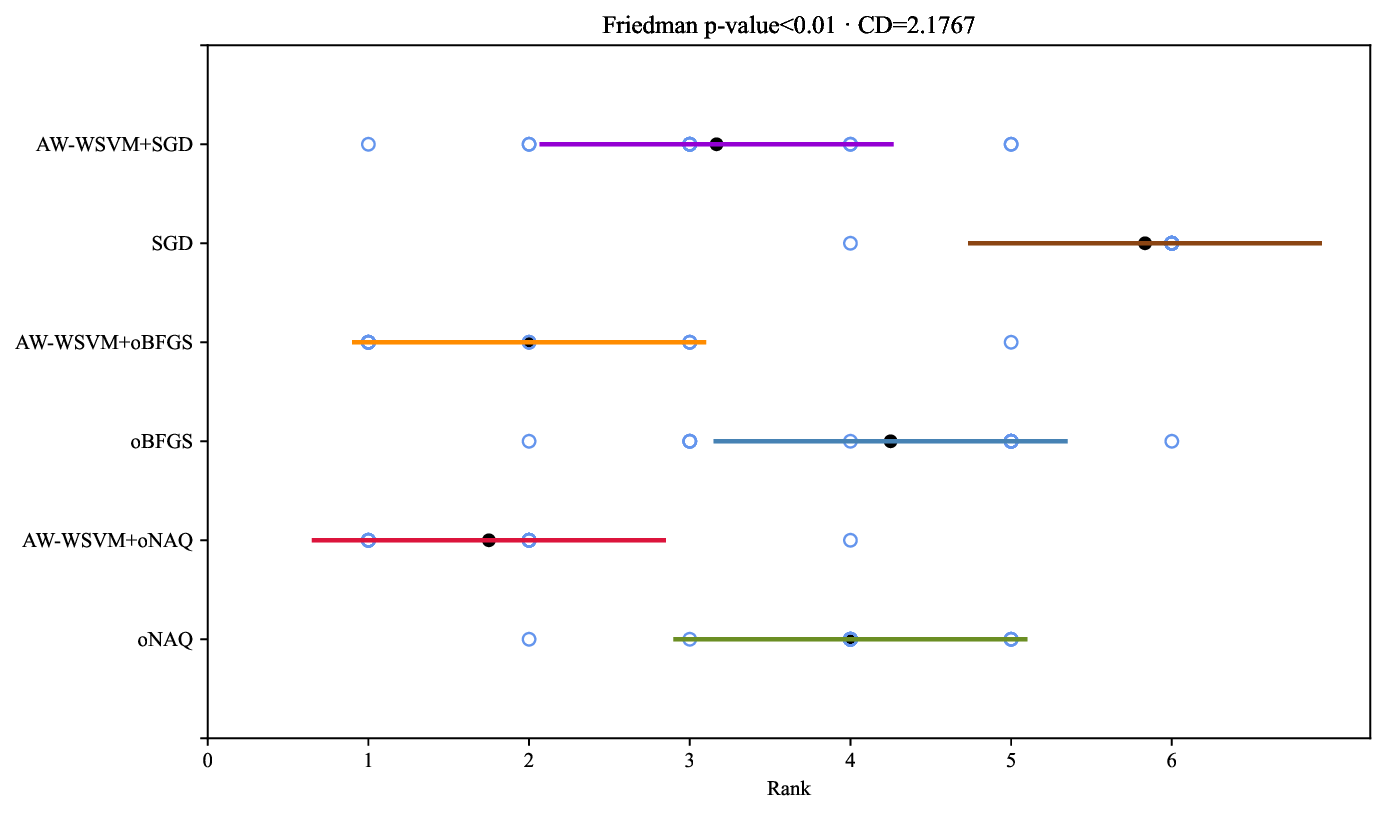}
	}
    \hfill % 是为了让多幅图在一行均匀分布（不加的效果是都挤在中间）
	\subcaptionbox{Classification F1-score\label{fig7(d)}}{
		\includegraphics[width=8cm]{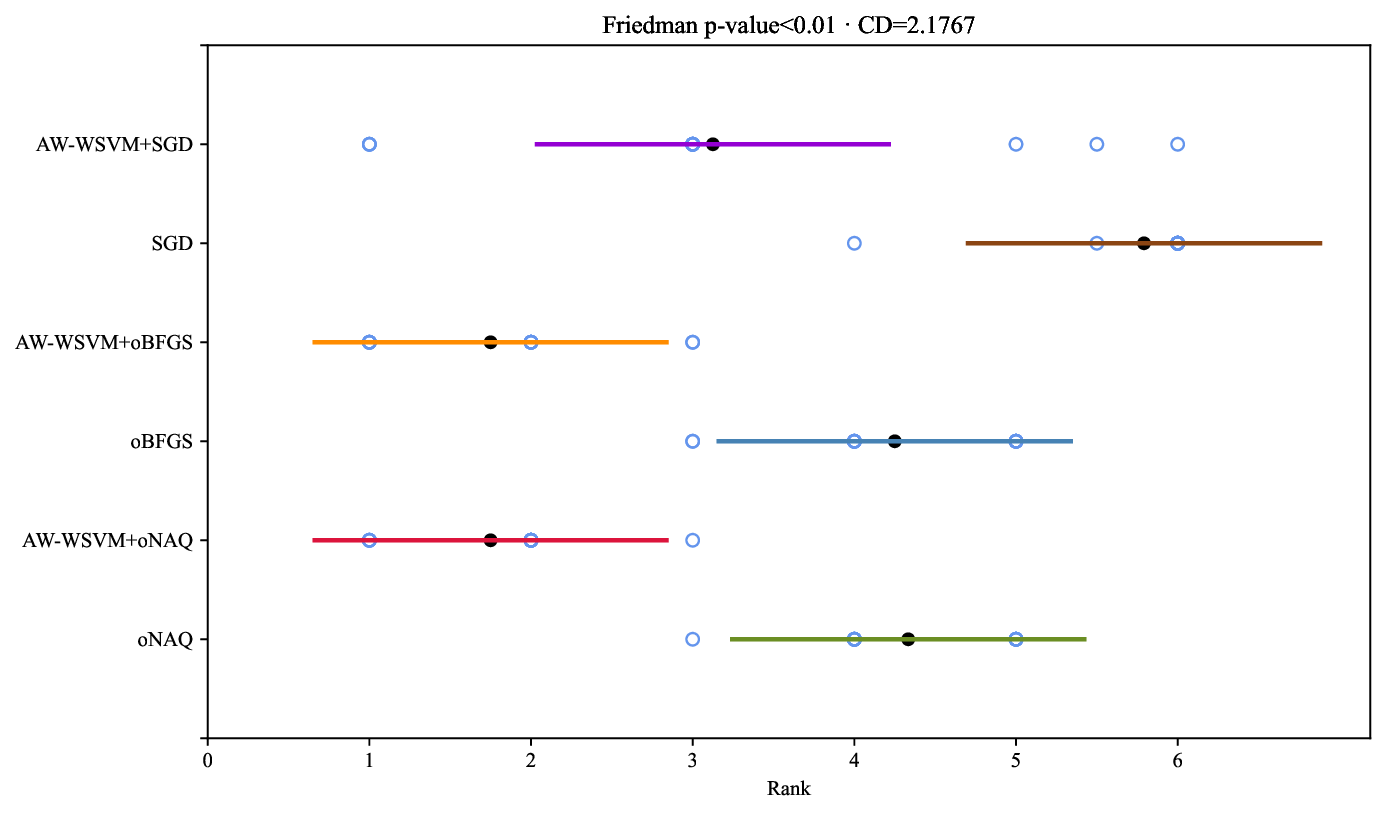}
	}
	\caption{Comparison of all classifiers against each other with the Nemenyi post-hoc test.}
	\label{fig7}
\end{figure}

Fig.\ref{fig7(a)}-\ref{fig7(d)} respectively report the analysis results of six classifiers in terms of test accuracy, G-mean, Recall and F1-score. The test accuracy of the standard SGD, oBFGS, and oNAQ algorithms does not exhibit any substantial variation, but there is a notable disparity between the proposed approach in this research. Furthermore, our analysis indicates that the integration of all three algorithms with AW-WSVM leads to enhanced algorithm performance. This improvement is evident from the overall higher ranking of the combined algorithms across all datasets, showcasing varied progress. The results indicate a negligible disparity in the classification accuracy across standard SGD, oBFGS, and oNAQ. Each of the three algorithms was examined independently. The performance of SGD was enhanced by using the weight update approach suggested in this research (AW-WSVM+SGD). Notably, there were notable disparities between the two algorithms when compared to the original SGD. The same applies to second-order approaches such as BFGS and NAQ.

\begin{table}[h]
\centering
\caption{Friedman ranking of these methods according to G-means on different datasets.}
\label{tab:9}
\setlength{\tabcolsep}{3.5mm}{
\begin{tabular}{l c c c c c c}
\toprule
Datasets&oNAQ&AW-WSVM+oNAQ&oBFGS&AW-WSVM+oBFGS&SGD&AW-WSVM+SGD\\
\midrule
A7a&6&5&4&3&2&1\\
A8a&6&3&5&4&2&1\\
A9a&4&2&5&3&6&1\\
Mushroom&4.5&1&4.5&2&5&3\\
Yeast&4&3&5&2&6&1\\
Ijcnn1&5&2&3&1&6&4\\
W1a&4&3&5&1&6&2\\
W2a&4&2&5&1&6&3\\
W3a&5&2&4&1&6&3\\
W4a&2&4&5&1&6&3\\
W5a&2&4&5&3&6&1\\
W6a&4&3&2&5&6&1\\
\hline
R&4.21&2.83&4.38&2.25&5.25&2.00\\
\bottomrule
\end{tabular}}
\end{table}

\begin{table}[h]
\centering
\caption{Friedman ranking of these methods according to Recalls on different datasets.}
\label{tab:10}
\setlength{\tabcolsep}{3.5mm}{
\begin{tabular}{l c c c c c c}
\toprule
Datasets&oNAQ&AW-WSVM+oNAQ&oBFGS&AW-WSVM+oBFGS&SGD&AW-WSVM+SGD\\
\midrule
A7a&2&1&5&3&6&4\\
A8a&5&4&3&1&6&2\\
A9a&5&1&3&2&6&4\\
Mushroom&4&1&2&3&6&5\\
Yeast&4&2&5&3&6&1\\
Ijcnn1&4&2&3&1&6&5\\
W1a&4&2&5&1&6&3\\
W2a&5&2&4&1&6&3\\
W3a&4&1&5&2&6&3\\
W4a&4&2&5&1&6&3\\
W5a&3&1&6&5&4&2\\
W6a&4&2&5&1&6&3\\
\hline
R&4.00&1.75&4.25&2.00&5.83&3.16\\
\bottomrule
\end{tabular}}
\end{table}

\begin{table}[h]
\centering
\caption{Friedman ranking of these methods according to F1-scores on different datasets.}
\label{tab:11}
\setlength{\tabcolsep}{3.5mm}{
\begin{tabular}{l c c c c c c}
\toprule
Datasets&oNAQ&AW-WSVM+oNAQ&oBFGS&AW-WSVM+oBFGS&SGD&AW-WSVM+SGD\\
\midrule
A7a&5&3&4&2&6&1\\
A8a&5&2&4&3&6&1\\
A9a&4&2&5&3&6&1\\
Mushroom&5&1&4&2&6&3\\
Yeast&4&2&3&1&5.5&5.5\\
Ijcnn1&5&2&3&1&4&6\\
W1a&3&1&4&2&6&5\\
W2a&5&2&4&1&6&3\\
W3a&4&1&5&2&6&3\\
W4a&4&2&5&1&6&3\\
W5a&4&1&5&2&6&3\\
W6a&4&2&5&1&6&3\\
\hline
R&4.33&1.75&4.25&1.75&5.79&3.12\\
\bottomrule
\end{tabular}}
\end{table}

\subsection{Experiments on Emotional Classification with Different Imbalance Ratio(IR)}
In this section, we collect the dataset of emotion classification, which comes from gitte\footnote{https://gitee.com/zhangdddong/toutiao-text-classfication-dataset.} (see Tabel \ref{tab:12}). The last column in the table indicates the imbalance ratio (IR). In daily life, different kinds of news data are always imbalanced, for example, entertainment news will account for more than science and technology news. 

In this section, the parameters are set as follows. The learning rate $\epsilon$ of SGD optimization algorithm is 0.1. The parameter $\tau$ in oBFGS and oNAQ optimization algorithm is set to 10 and $\gamma$ is set to 0.2. The momentum coefficient $\mu$ to 0.1. The highest limit of iterations is set to 500 and batchsize of all algorithms is set to 256.

Table \ref{tab:13} and Table \ref{tab:14} record the test accuracy and G-means after 100 iterations, respectively, in which bold indicates the best results compared with the two. It is well known that G-means can be used as an effective indicator to measure the quality of the model when the data is unbalanced. Therefore, we can draw the conclusion that our framework, AW-WSVM, will effectively improve the performance of the standard stochastic optimization algorithm, no matter whether the IR is high or low.

\begin{figure}[h]
	\subcaptionbox{102vs100(SGD)}{
		\includegraphics[width=5.2cm]{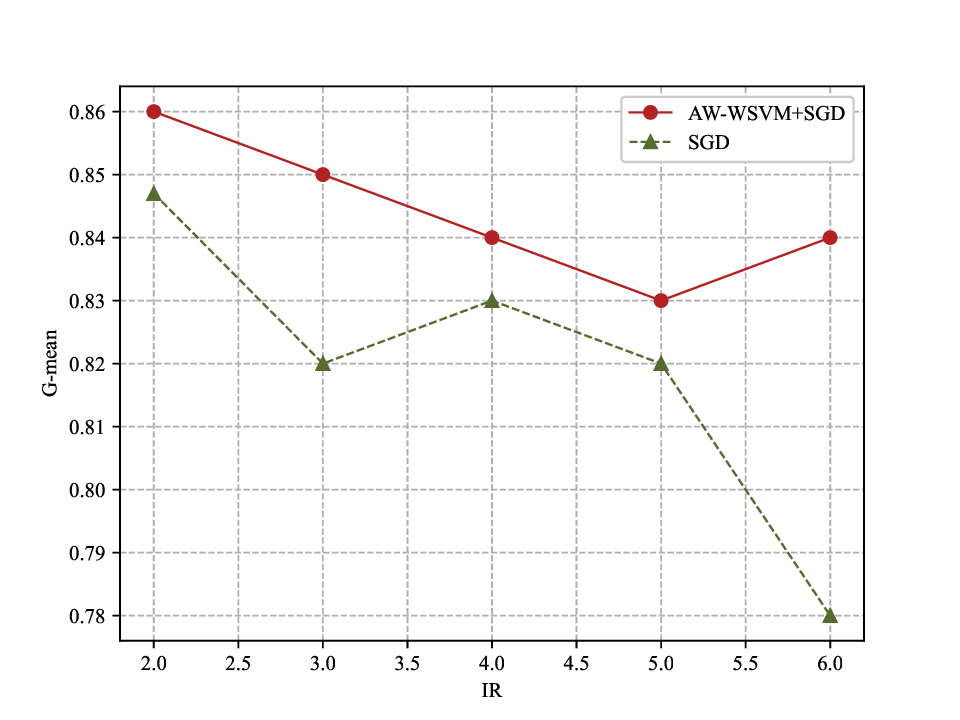}
	}
	\hfill % 是为了让多幅图在一行均匀分布（不加的效果是都挤在中间）
	\subcaptionbox{102vs100(oBFGS)}{
		\includegraphics[width=5.2cm]{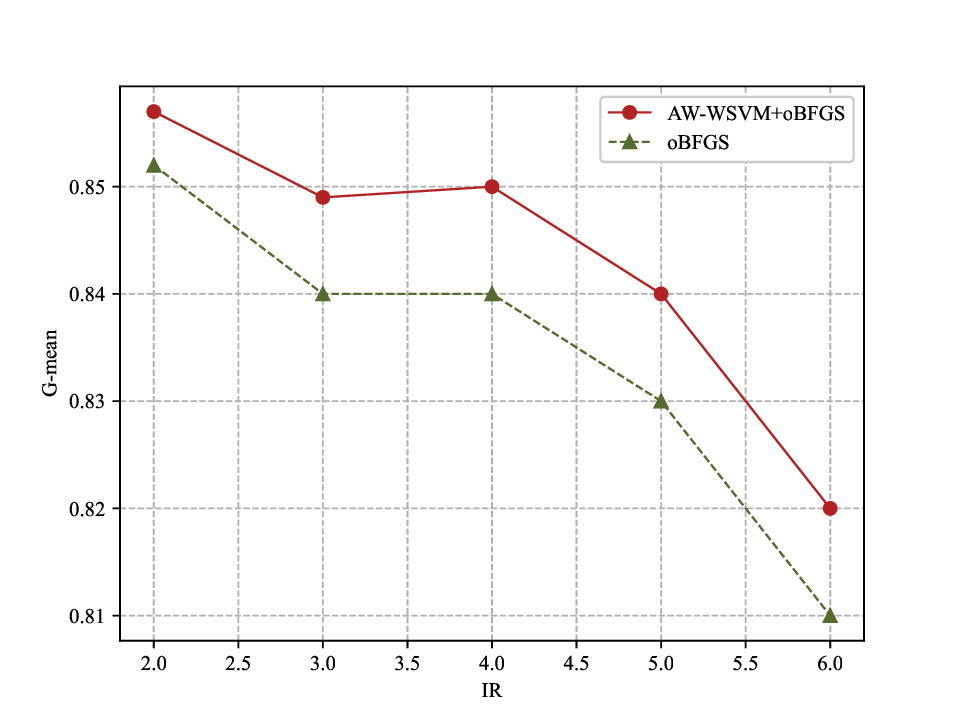}
	}
 	\hfill % 是为了让多幅图在一行均匀分布（不加的效果是都挤在中间）
	\subcaptionbox{102vs100(oNAQ)}{
		\includegraphics[width=5.2cm]{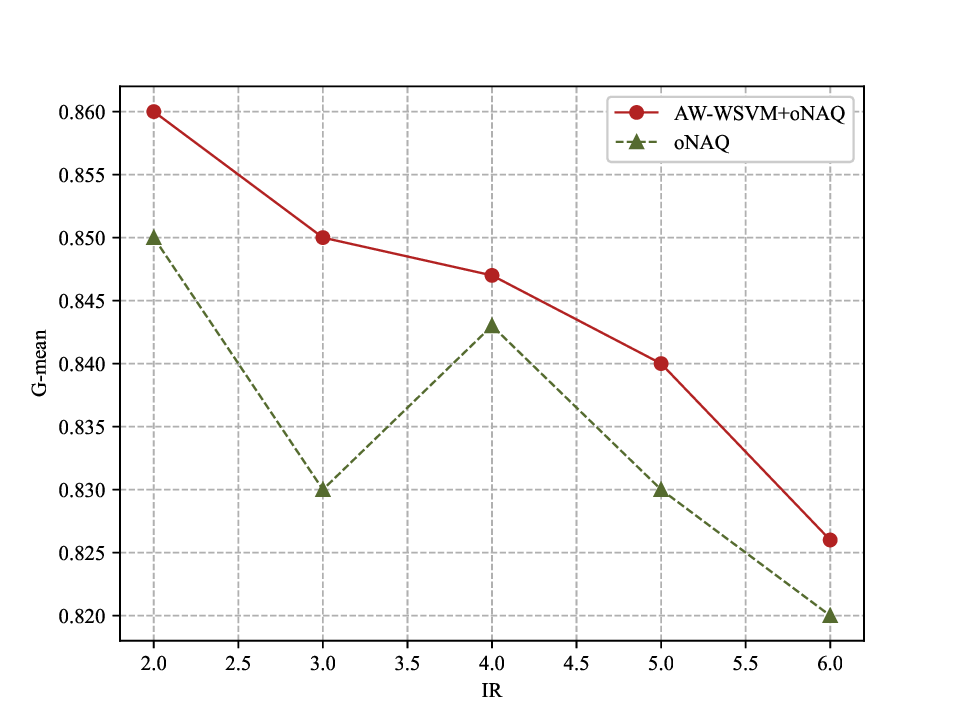}
	}
   \hfill
    \subcaptionbox{109vs100(SGD)}{
		\includegraphics[width=5.2cm]{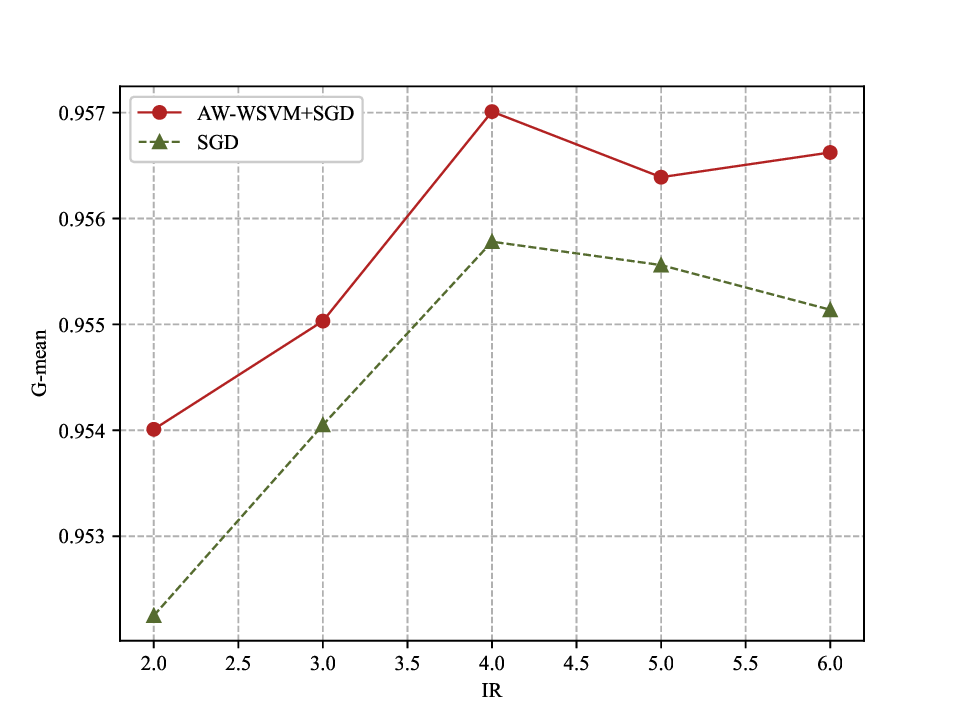}
	}
	\hfill % 是为了让多幅图在一行均匀分布（不加的效果是都挤在中间）
	\subcaptionbox{109vs100(oBFGS)}{
		\includegraphics[width=5.2cm]{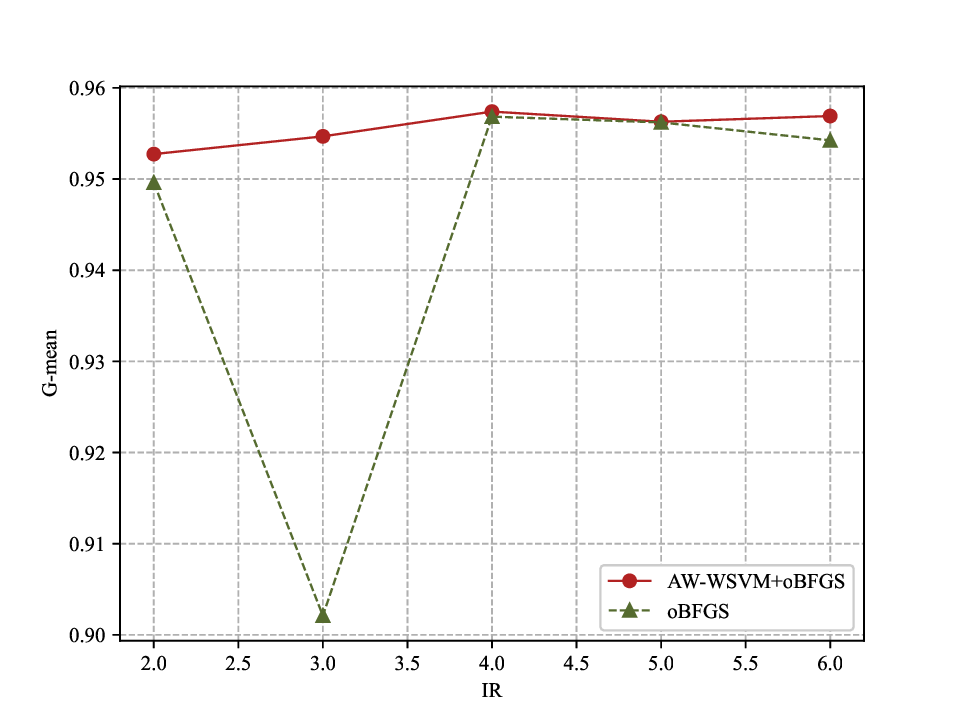}
	}
 	\hfill % 是为了让多幅图在一行均匀分布（不加的效果是都挤在中间）
	\subcaptionbox{109vs100(oNAQ)}{
		\includegraphics[width=5.2cm]{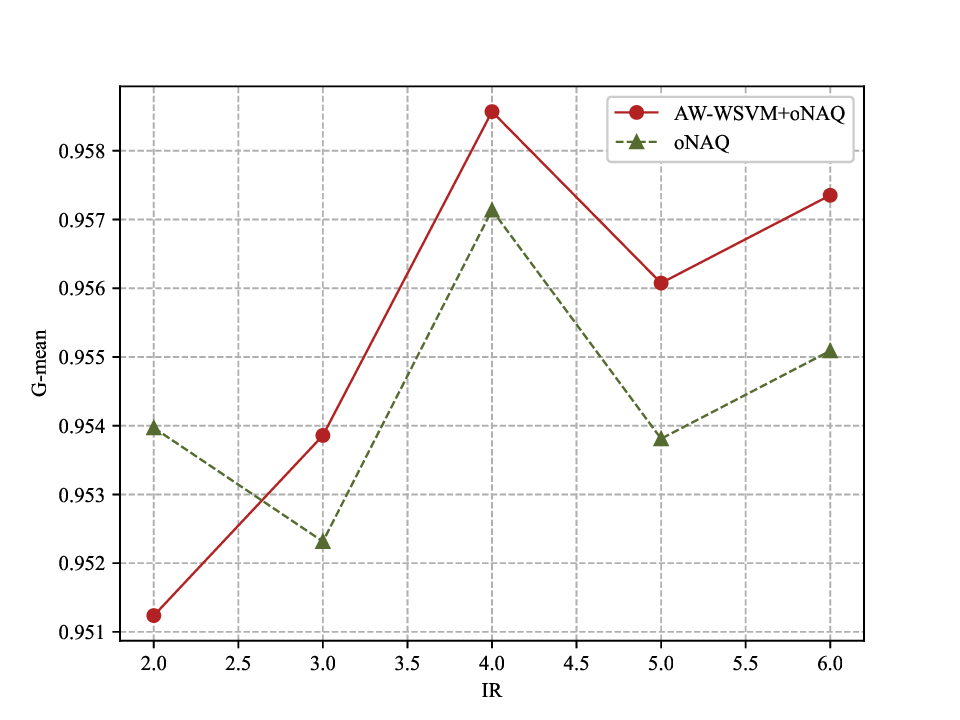}
	}
	\caption{Comparisons on different IRs.}
	\label{fig8}
\end{figure}

In order to further illustrate the effectiveness of the proposed AW-WSVM framework, we process the same data and get G-means under different IRs\cite{43}. Taking 102vs100 and 109vs100 as examples, we plot the results in Fig\ref{fig8}. It is not difficult to find that our proposed method performs best in both low IR and high IR.

\begin{table}[h]
\centering
\caption{The description of the emotional datasets.}
\label{tab:12}
\setlength{\tabcolsep}{4mm}{
\begin{tabular}{l l l l l}
\toprule
Datasets&Samples&Features&Classes&IR\\
\midrule
102\_vs\_100&45669&12&2&6.28\\
109\_vs\_116&70843&12&2&1.42\\
109\_vs\_100&47816&12&2&6.62\\
107\_vs\_106&53457&12&2&2.02\\
103\_vs\_115&56890&12&2&1.94\\
101\_vs\_100&34304&12&2&4.47\\
% Toutiao&Gitee&70843&128&2\\
\bottomrule
\end{tabular}}
\end{table}

\section{Conclusion}
We proposes a novel generalized framework, referred to as AW-WSVM, for SVM. The purpose of its development was to address the complexities of real-world applications by effectively handling uncertain information. It ensures that training remains effective even when dealing with challenging and extensive datasets, without compromising on speed. 

Considering the massive data, we propose an Adaptive Weight function (AW function), through which the weights of samples can be dynamically updated. Our analysis demonstrates that the weights generated by the AW function obey a probability distribution in each class. We propose a new soft-margin weighted SVM and provide a more concise matrix expression, so that any stochastic optimization algorithm can iterate through the matrix, saving computational space. Since minority is extremely sensitive to noise, we propose an effective noise filtering method based on nearest neighbors, which improves the robustness of the model. Subsequently, we can allocate greater weights to samples that are in close proximity to the decision hyperplane, while assigning lower weights to the other samples. Our allocation technique is appropriate since the support vectors, which are placed near the decision hyperplane, correspond with the concept that samples close to the decision plane have higher influence. 

\begin{table}
\centering
\caption{Accuracy of different methods on emotional datasets.}
\label{tab:13}
\setlength{\tabcolsep}{3.5mm}{
\begin{tabular}{l c c c c c c}
\toprule
Datasets&oNAQ&AW-WSVM+oNAQ&oBFGS&AW-WSVM+oBFGS&SGD&AW-WSVM+SGD\\
\midrule
102\_vs\_100&0.9541&\textbf{0.9558}&0.9533&\textbf{0.9564}&0.9465&\textbf{0.9580}\\
109\_vs\_116&0.9125&\textbf{0.9189}&0.9123&\textbf{0.9197}&0.9058&\textbf{0.9186}\\
109\_vs\_100&0.9895&\textbf{0.9902}&0.9902&\textbf{0.9907}&0.9894&\textbf{0.9899}\\
107\_vs\_106&0.9700&\textbf{0.9734}&0.9703&\textbf{0.9737}&0.9664&\textbf{0.9724}\\
103\_vs\_115&0.9825&\textbf{0.9832}&0.9819&\textbf{0.9833}&0.9822&\textbf{0.9832}\\
101\_vs\_100&0.9721&\textbf{0.9735}&0.9726&\textbf{0.9733}&0.9708&\textbf{0.9730}\\
\bottomrule
\end{tabular}}
\end{table}

\begin{table}
\centering
\caption{G-mean of different methods on emotional datasets.}
\label{tab:14}
\setlength{\tabcolsep}{3.5mm}{
\begin{tabular}{l c c c c c c}
\toprule
Datasets&oNAQ&AW-WSVM+oNAQ&oBFGS&AW-WSVM+oBFGS&SGD&AW-WSVM+SGD\\
\midrule
102\_vs\_100&0.8203&\textbf{0.8260}&0.8169&\textbf{0.8278}&0.7817&\textbf{0.8353}\\
109\_vs\_116&0.8340&\textbf{0.8454}&0.8335&\textbf{0.8468}&0.8228&\textbf{0.8447}\\
109\_vs\_100&0.9533&\textbf{0.9561}&0.9563&\textbf{0.9585}&0.9530&\textbf{0.9550}\\
107\_vs\_106&0.9346&\textbf{0.9417}&0.9352&\textbf{0.9424}&0.9274&\textbf{0.9397}\\
103\_vs\_115&0.9620&\textbf{0.9635}&0.9607&\textbf{0.9636}&0.9613&\textbf{0.9634}\\
101\_vs\_100&0.9136&\textbf{0.9147}&0.9117&\textbf{0.9143}&0.9063&\textbf{0.9121}\\
\bottomrule
\end{tabular}}
\end{table}

The proposed framework, AW-WSVM, was evaluated on a total of 12 different datasets sourced from UCI repositories and LIBSVM repositories. The experiments conducted confirm that the proposed framework AW-WSVM surpasses the conventional first-order approaches SGD, second-order methods oBFGS, and oNAQ. Furthermore, it exhibited superior performance in terms of classification accuracy, g-mean, recall and F1-score.  In addition, we also conducted statistical experiments on these six classifiers. Friedman tests and Nemenyi post hoc tests demonstrate that the standard algorithms SGD, oBFGS, and oNAQ all perform better within our proposed framework. We not only do experiments on standard datasets, but also apply the framework to emotion classification datasets. From the G-means and other indicators, the framework proposed in this paper can effectively improve the performance of standard optimization algorithms, no matter whether the imbalance index is high or low. Moreover, the core of the proposed generalized framework, AW-WSVM, is to look for data samples near the classified hyperplane. Consequently, it may be effectively integrated with any soft-margin SVM solution techniques.

% \bmhead{Acknowledgements}
\section*{Acknowledgement}
This work is supported by the National Natural Science Foundation of China (Grant No.61703088), the Fundamental Research Funds for the Central Universities (Grant No.N2105009)

\section*{Declaration of Generative AI and AI-assisted technologies in the writing process}

During the preparation of this work, the authors used ChatGPT in order to translate and revise some parts of the paper. After using this tool, the authors reviewed and edited the content as needed and took full responsibility for the content of the publication.

\makeatletter
\renewcommand\@biblabel[1]{#1.}
\makeatother

\bibliography{sn-bibliography}% common bib file
\end{CJK}
\end{document}